\documentclass[lettersize,journal]{IEEEtran}
\usepackage{amsmath,amsfonts,amssymb}
\usepackage{algorithm}
\usepackage{array}
\usepackage[caption=false,font=normalsize,labelfont=sf,textfont=sf]{subfig}
\usepackage{textcomp}
\usepackage{bbm}
\usepackage{textcomp}
\usepackage{float}
\usepackage{url}
\usepackage[dvipsnames]{xcolor}
\usepackage{verbatim}
\usepackage{graphicx}
\usepackage{cite}
\usepackage{algpseudocode}
\usepackage{transparent}
\usepackage{bbm}
\usepackage[justification=centering]{caption}
\usepackage{setspace}
\usepackage{parskip}
\usepackage{stfloats}
\usepackage{tikz}

\begin{document}

\title{Pedestrian Trajectory Forecasting Using Deep Ensembles Under Sensing Uncertainty}

\author{Anshul Nayak, Azim Eskandarian, Zachary Doerzaph, Prasenjit Ghorai
\thanks{Autonomous Systems and Intelligent Machines lab, Virginia Tech.\\
 }
}



\maketitle

\begin{abstract}
One of the fundamental challenges in the prediction of dynamic  agents is robustness. Usually, most predictions are deterministic estimates of future states which are over-confident and prone to error. Recently, few works have addressed capturing uncertainty during forecasting of future states. However, these probabilistic estimation methods fail to account for the upstream noise in perception data during tracking. Sensors always have noise and state estimation becomes even more difficult under adverse weather conditions and occlusion. Traditionally, Bayes filters have been used to fuse information from noisy sensors to update states with associated belief. But, they fail to address non-linearities and long-term predictions. Therefore, we propose an end-to-end estimator that can take noisy sensor measurements and make robust future state predictions with uncertainty bounds while simultaneously taking into consideration the upstream perceptual uncertainty.  For the current research, we consider an encoder-decoder based deep ensemble network for capturing both perception and predictive uncertainty simultaneously. We compared the current model to other approximate Bayesian inference methods. Overall, deep ensembles provided more robust predictions and the consideration of upstream uncertainty further increased the estimation accuracy for the model.

\end{abstract}

\begin{IEEEkeywords}
Uncertainty quantification, Bayesian Inference, Deep Ensembles, MC Dropout
\end{IEEEkeywords}

\section{INTRODUCTION}

 Most of the prediction algorithms output  deterministic estimates of future states from raw sensor data \cite{Houenou}. Deterministic predictions are over-confident and prone to error. Therefore, it is important to make probabilistic predictions of future states to improve robustness downstream, especially for uncertainty-aware planning \cite{Kahn}\cite{Xihui}.
 Past research has tried to address the issue by developing probabilistic  methods for prediction \cite{Nayak, Jiachen, Wiest}. One popular approach is to use Bayesian  neural network (BNN)  to capture uncertainty in both classification and regression problems \cite{Jospin}. However, exact Bayesian inference is computationally challenging due to a large number of model parameters. Therefore, approximate inference methods like Monte Carlo dropout \cite{Gal} and deep ensembles \cite{Lakshminarayanan} have been developed which can output  probabilistic predictions approximating  the posterior without making significant changes to the neural network (NN) architecture.  The deep ensemble model has a network of independently trained neural networks with each network having a random initialization.  Each network outputs probabilistic predictions based on a sophisticated loss function and the predictions are averaged over all the networks to obtain the predictive mean and variance assuming Gaussian posterior distribution. Meanwhile, the MC dropout method introduces dropout \cite{Srivastava} layers during training and inference to capture predictive uncertainty. During inference, weights are randomly dropped to generate a distribution of outputs rather than deterministic predictions.   The  predictive uncertainty accounts for noise in the output data called aleatoric as well as the  variation in predictions by  the neural network model  known as epistemic uncertainty  \cite{Valdenegro-Toro}. Feng et.al \cite{Feng} designed a NN architecture for Lidar 3D vehicle detection and localization capturing aleatroic and epistemic uncertainty during predictions.

 \begin{figure*}[h!]
   \centering      
   \includegraphics[width=1\textwidth]{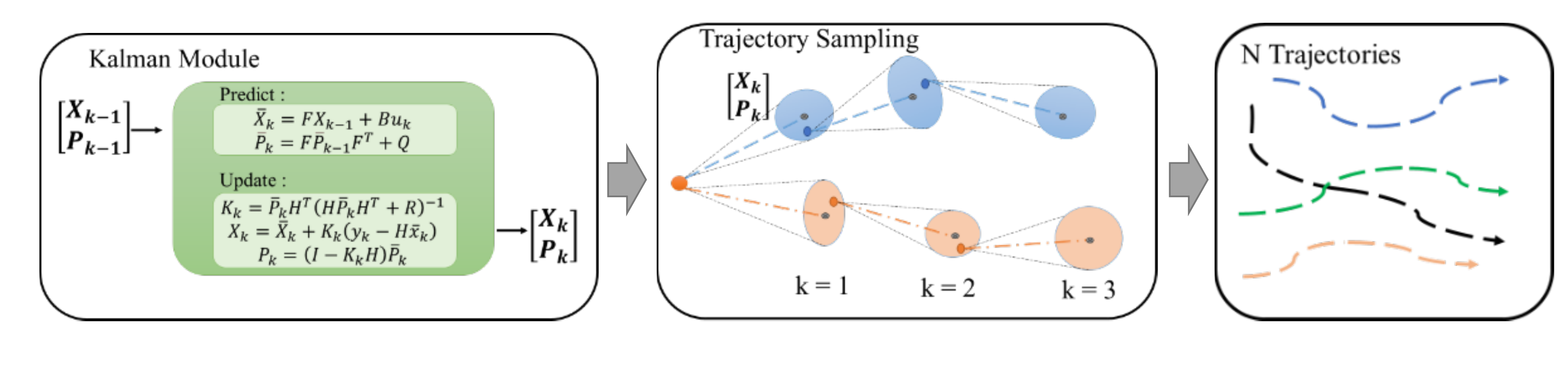} 
 \caption{\textbf{Model}: The Kalman filter module updates the state and covariance at each step. \textbf{Trajectory Sampling}: Conditional Trajectory Sampling propagates an initial state based on system dynamics to generate a trajectory. TS is called recursively to generate a distribution of trajectories.}
 \label{fig:figure1}
 \end{figure*}
 
 Although, the prediction model provides probabilistic outputs for future states, the model itself considers deterministic states during training and prediction. In the context of prediction, the model assumes that the input states as observed by the sensors are deterministic and makes predictions based on these deterministic state inputs. However,  the sensors are inherently noisy and can not accurately estimate the state of an object. Further,  state estimation becomes even more uncertain when coupled with adverse weather or  occlusion \cite{Yoneda}\cite{Wolfram}. Therefore, having deterministic state estimation may not be correct, and capturing the perception uncertainty associated with each state is necessary for robust predictions downstream. The main idea of the current paper is to investigate why incorporating and propagating perceptual uncertainty into the prediction pipeline is necessary. Traditionally, Bayes filters have been used to capture the state and its associated covariance during tracking \cite{Prevost}\cite{Breitenstein}. A simple Bayes filter like Kalman Filter (KF) recursively updates the state and covariance at each step by fusing raw noisy sensor measurements with the prior computed using a motion model. But, the KF can reliably estimate only the states for which it has measurements while our problem  requires the model to accurately learn and predict the covariance associated with  future states.  Deep neural networks have been utilized to estimate covariance from raw sensor data \cite{DICE}. The authors  learned the representation for measurement model by minimizing the loss between ground truth and raw sensor measurement. Similarly, Bertoni et.al \cite{Bertoni} captured 3D localization uncertainty during tracking through a loss function based on Laplace distribution. They used MC dropout and captured both aleatoric and epistemic uncertainty during state localization using monocular RGB images.  Recently, Rebecca et.al \cite{Rebecca} modelled multivariate uncertainty for regression problems by training a NN end-to-end through a KF.  Our model draws inspiration from previous work and has a simple encoder-decoder architecture that  learns the  KF covariance  by minimising the MSE loss between model and ground truth covariance measurements via supervised regression. The learned NN model is able  to estimate covariance for future states capturing perceptual uncertainty. Once the model learns to estimate covariance,  we propagate the sensing uncertainty into prediction pipeline. 

 We design the current NN model as an end-to-end estimator that can simultaneously predict the perception and prediction uncertainty associated with future states. In the past,  end-to-end approaches such as FAF\cite{Wenjie} projected Lidar points into the bird's eye view (BEV) grid generating predictions by inferring detection multiple times in future.   Further,  PTP \cite{Weng} unified the Multi-object tracking (MOT) and prediction under one framework. However, they used  generative modeling for trajectory distribution which is not particularly efficient in  incorporating the propagation of  perception uncertainty into prediction.  More recently, Pavone et.al \cite{Ivanovic} showed the importance of  propagating state uncertainty  and leveraged upon the idea of penalizing the loss function to  encode state uncertainty. Our approach in a way combines both the notion of end-to-end tracking and prediction while simultaneously estimating the perceptual uncertainty for robust probabilistic trajectory predictions. We designed our model on a sophisticated loss function that minimises mean-squared error (MSE) and negative-log-likelihood (NLL) loss simultaneously to perform robust end-to-end predictions while estimating perceptual uncertainty. The model learns to  perform state covariance estimation by minimising the MSE loss with KF ground truth covariance. Meanwhile, the predictive uncertainty is captured by minimising the NLL loss using a deep ensemble model. Overall, our  end-to-end NN model can take raw sensory inputs with measurement noise and make robust probabilistic predictions of future states downstream without ignoring the upstream perceptual uncertainty.

\textbf{Contributions.} Our key contributions are as follows.  We propose  a simple end-to-end NN model that can capture both perceptual and predictive uncertainty for robust state prediction. Secondly, we show the essence of using deep ensembles for predictive uncertainty. Finally, we also show how incorporating state uncertainty into prediction pipeline improved overall robustness and compared the deep ensemble model with  MC dropout on publicly available pedestrian datasets. Further, we performed  offline experiments with  the trained model to understand the out-of-distribution prediction accuracy.

\begin{table}[ht]
\caption{NOTATION }
\begin{center}
\begin{tabular}{||c c ||} 
\hline
 $\mathbf{X}$ & State   \\ 
  \hline
  $F: R^{n}\rightarrow  R^{n}$ & State Transition matrix  \\

 \hline
 $z \in R^{n}$ & Raw measurement  \\
 \hline
  $P \in R^{n x n}$ & Posterior covariance  \\
 \hline
$K \in R^{k x n}$ & Kalman Gain \\
 \hline
 H : $R^{n}\rightarrow  R^{k}$  & Observation matrix  \\
 \hline

 $Q \in R^{n x n}$ & Process Noise covariance\\ [1ex] 
\hline
 $R \in R^{n x n}$ & Measurement covariance  \\ [1ex] 
 \hline

\end{tabular}

\end{center}
 \label{table:table1}
\end{table}

\section{PROPOSED METHOD}\label{sec:methods}

  Our method  is two-fold; first  the neural network model approximates  a Bayes filter to predict the sensing uncertainty at each future state.   Secondly, a prediction  model considers the upstream sensing uncertainty and makes robust future state predictions downstream.  we use Deep Ensembles for quantifying predictive uncertainty while sensing covariance estimation is carried out using a simple Kalman filter.

\begin{figure*}[ht]
   \centering      
   \includegraphics[width=1\textwidth]{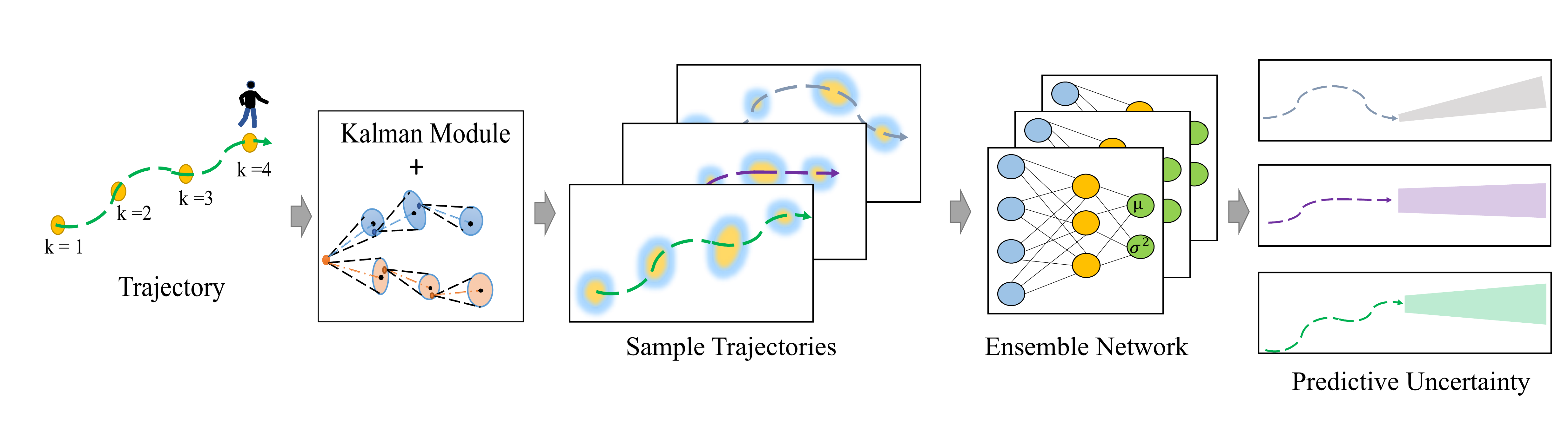} 
 \caption{\textbf{Deep Ensemble:} NN model takes a single trajectory and generates a distribution of input trajectories using KF and Trajectory Sampling. Each trajectory is trained through am independent NN within the ensemble to generate predictive uncertainty.}
 \label{fig:figure2}
 \end{figure*}

\subsection{Covariance Estimation using Bayes Filter}

In this section, we present an approach for estimating the state and covariance of a Bayes filter using a neural network (NN). Bayes filters are  used  to account for the uncertainty during sensing and localization with popular filters such as the KF, Extended Kalman filter, and Particle filter. These filters estimate the states of a system as well as  their associated beliefs, which are updated by fusing information from raw sensor measurements. The belief, which characterizes the state uncertainty, can have either uniform variance, known as homoskedastic, or heteroscedastic noise. Capturing this upstream uncertainty during  sensing data is critical for subsequent robust predictions. Therefore, the current objective is to design a NN model that can learn the  heteroscedastic measurement covariance of a Bayes filter.
We approach the problem of state estimation as a 2D object  tracking problem, using the KF. 

The KF module consists of two steps; prediction and update step (Figure \ref{fig:figure1}). The prediction step takes the previous state $\mathbf{X_{k-1}}$ and covariance $\mathbf{P_{k-1}}$ and computes the prior distribution based on the constant velocity motion model \eqref{eq:predction}.    

\begin{equation*}
    {\mathbf{\bar{X}_{k}}} = \mathbf{F}\mathbf{X_{k-1}} + \mathbf{B}u_{k} 
\end{equation*}
\begin{equation}
    \mathbf{\bar{P}_{k}} = \mathbf{F}\mathbf{P_{k-1}}\mathbf{F}^{T} + \mathbf{Q}
    \label{eq:predction}
\end{equation}

Here, $\mathbf{F}$ is the state transition matrix, and $u_{k}$ is the control input. For the tracking problem, the control input has no significance. Further,  $\mathbf{Q}$ represents the  process noise covariance.

We model the innovation, $z_{k}$ based on the difference between actual and predicted measurement of the state
\begin{equation*}
    z_{k} = y_{k} - \mathbf{H}{\mathbf{\bar{X}_{k}}}
    \end{equation*}
    which has covariance, 
\begin{equation*}
    \mathbf{S_{t}} = \mathbf{R} +(\mathbf{H} \mathbf{\bar{P}_{k}}\mathbf{H}^{T})
\end{equation*}
It is the sum of measurement noise covariance and predicted state covariance, $\mathbf{\bar{P}_{k}}$. R represents the covariance matrix associated with measurement noise.
\begin{equation*}
    \mathbf{K_{k}} =  \mathbf{\bar{P}_{k}}\mathbf{H}^{T}(\mathbf{H}{ \mathbf{\bar{P}_{k}}}\mathbf{H}^{T} + \mathbf{R})^{-1}
\end{equation*}
 The state, $\mathbf{{X}_{k}}$ and associated covariance, ${\mathbf{P_{k}}}$  are updated at each step using the Kalman gain, $ \mathbf{K_{k}}$ which resembles a weighting factor between predicted (prior) state and actual measurement (likelihood) of the state.
\begin{equation*}
   \mathbf{{X}_{k}} = \mathbf{\bar{X}_{k}} + \mathbf{K_{k}}y_{k}
    \end{equation*}
\begin{equation}
    {\mathbf{P_{k}}} = \mathbf{(I - K_{k}H)}\mathbf{\bar{P}_{k}}
    \label{eqn:states_post}
\end{equation}

The posterior states, $\mathbf{{X}_{k}}, \mathbf{P_{k}} $ \eqref{eqn:states_post} are updated recursively at each step by fusing the actual noisy measurement from sensors and predicted measurement using the motion model.

\subsection{Conditional Trajectory Sampling}

Usually, trajectories can be randomly sampled from the posterior distribution of states obtained using the KF. However, random sampling generate non-smooth and infeasible trajectories especially with high sensor noise. To address this in-feasibility, we  propose conditional trajectory sampling (CTS) where we propagate an initial sampled  point using a dynamics model  \cite{Chua} and then resample a new point from the adjacent posterior distribution. This process is repeated recursively till a trajectory is generated (Figure \ref{fig:figure1}). In this way, the generated trajectories adhere to  system  dynamics and are not random.   

 The CTS technique is used to sample from the  posterior distribution of the KF state covariance at each time step.
  We randomly sample the initial state as $x_{0}$  and then propagate  the initial  state based on some motion model as a Markovian process $x_{t} \sim P(x_{t}|x_{t-1}, a)$ \cite{Rose}. The prior distribution, $P(x_{t-1})$ is propagated based on some action, \textit{a}. This   transition predicts the the likelihood of  next state based on the constant-velocity motion model. The Process is repeated recursively based on the Markovian dynamics to generate a trajectory. Further, we repeat the trajectory generation process according to a particular bootstrap, i $\in$ \{1,...,M\}, where M represents the number of trajectories. The ensemble of generated trajectories roughly represent the state uncertainty.  For the current research, trajectory sampling can be considered as a data augmentation step for the training of the neural network to capture this perceptual uncertainty.  Another benefit of CTS is that each sampled trajectory  within the distribution  can be fed into the independent NN of an ensemble model to capture total predictive uncertainty. Details of our implementation are discussed in sec.\ref{sec:deep_ensembles} and Figure \ref{fig:figure2}.   

     
 
\subsection{Uncertainty Estimation}

We usually treat trajectory prediction as a regression problem. For regression problems with deterministic predictions, the NNs output a single value say $\mu(x)$. It is estimated by minimizing the mean squared error on the training set, MSE = $\sum_{n=1}^{N} (y_{n} - \mu({x_{n}}))^{2} $. However, the outputs, $y_{n}$ are point estimates and do not contain any information on associated uncertainty. To capture the  uncertainty, we assume the outputs are sampled from a Gaussian distribution such that the final layer outputs two values,  predicted mean, $\mu({x})$ and variance, $\sigma^{2}(x)$ of the distribution. The variance, $\sigma^{2}(x)$ of a NN model can be obtained using the Gaussian negative log-likelihood loss (NLL) on training samples with input $x_{n}$ and output $y_{n}$ as:
\begin{equation}
    -logP(y_{n}|x_{n}) = \frac{log\,\sigma^2(x_{n})}{2} + \frac{(y_{n} - \mu({x_{n}}))^{2}}{2\sigma^2(x_{n})} + constant
\end{equation}
$\sigma(x)$ represents the model's noise observation parameter - showing the amount of noise present in the model's outputs. However, the standard NLL strongly depends on predictive variance, $\sigma(x)$ and scales down the gradient for ill-predicted points \cite{Valdenegro-Toro}. Hence, an alternative loss function called as the $\beta$-exponentiated negative log-likelihood loss ($\beta$-NLL) \cite{Seitzer} has been used to minimize loss. 
\begin{equation}
    {L_{\beta-NLL}} = -logP(y_{n}|x_{n}) \, stop(\sigma^{2\beta})
    \label{beta_NLL}
\end{equation}
$\beta$ controls the dependency of gradients on predictive variance while stop() is the stop gradient operation that prevents the gradients from flowing.  $\beta$ = 0 represents the standard NLL loss. Meanwhile, $\beta$ =1 completely removes the dependency of gradients on variance, $\sigma(x)$
and treats the loss function as standard mean-squared error (MSE). The $\beta$-NLL loss function allows us the flexibility to switch between NLL and MSE loss function.

\subsubsection*{\textbf{Deep Ensembles}}\label{sec:deep_ensembles} \hfill\\ 
Deep ensemble is an approximate Bayesian inference method that can capture   predictive uncertainty  during forecasting. It is  simple and scalable compared to Bayesian NNs. As the name suggests, an ensemble network consists of a series of NNs which are different from one another due to random initialization.
Let, M denote the number of NNs present within the ensemble. Then,  $\mu_{i}(x)$ and $\sigma_{i}(x)$ represent the mean and variance of a single NN indexed i $\in$ [1,...,M]. Balaji et.al \cite{Lakshminarayanan} treated the ensemble as a uniformly-weighted mixture model and combine the predictions into a single Gaussian mixture distribution $p(y|x)$ using:
    \begin{equation}
        p(y|x) \sim N(\mu_{i}(x), \sigma^2_{i}(x))
    \end{equation}
    And for ease of estimating predictive probabilities, they further approximated the ensemble prediction as a Gaussian whose mean and variance correspond to the respective mean and variance of the mixture model.
    \begin{equation}
    \mu_{*}(x) = M^{-1}\sum_{i}{} \mu_{i} (x)
    \end{equation}
    \begin{equation}
    \sigma_{*}^{2}(x) = M^{-1}\sum_{i}{} (\sigma^{2}_{i} (x) + \mu^{2}_{i} (x)) - \mu^{2}_{*}(x)
    \label{predictive_var}
    \end{equation}

\subsubsection*{\textbf{Dropout as Bayesian approximation}}  \label{sec:Dropout}\hfill\\ We compare the performance of other approximate Bayesian inference methods used for uncertainty quantification with deep ensembles \cite{Lakshminarayanan}.
One such approach focuses on 
 dropout \cite{Srivastava}  to capture the total predictive uncertainty. The key notion is to randomly drop weights  during both training and inference. Concisely, we can formulate the MC Dropout algorithm as,

 \begin{equation*}
 \begin{aligned}
 \text{for b = 1:B }\\
    & e_{(b)}^{*} = \textit{VariationalDropout}(g(x^{*}), p)\\
    & y_{(b)}^{*} = 
    \textit{Dropout}
    (h(e^{*}), p)\\
    \text{end for}\\
\end{aligned}
\end{equation*}
 Provided the input data $x^{*}$, an encoder-decoder network $g(.)$, prediction network $h(.)$, dropout probability $p$, and number of iterations B, we  train the encoder-decoder model, $e = g(.)$   with  dropout, $p$. Further, during inference, for the same input, $x^{*}$, the prediction network, $h(.)$ is inferred by randomly dropping weights to generate a distribution of B outputs. \cite{Gal} showed the output distribution approximates a Bayesian NN without the additional complexity. The mean and variance of the predicted samples are presented below. 
\begin{equation*}
     \\
    \hat{y}_{mc}^{*} = \frac{1}{B}\sum_{b=1}^{B} \hat{y}_{(b)}^{*}
\end{equation*}

\begin{equation}
    \eta_{1}^{2} = \frac{1}{B}\sum_{b=1}^{B} (\hat{y}_{(b)}^{*} - \hat{y}_{mc}^{*})^{2}
\end{equation}

\subsection{Uncertainty Disentanglement}

Total predictive variance \eqref{predictive_var} can be disentangled into aleatoric uncertainty, associated with the inherent noise of the data, and epistemic uncertainty accounting for uncertainty in model  predictions\cite{Valdenegro-Toro}. 
\begin{equation}
\begin{aligned}
     \sigma_{*}^{2}(x) &= M^{-1}\sum_{i}{} \sigma^{2}_{i}(x) &+  M^{-1}\sum_{i}\mu^{2}_{i} (x) - \mu^{2}_{*}(x)\\ 
     &=  \mathbb{E}_{i} [\sigma^{2}_{i} (x)] &+ 
     \mathbb{E}_{i} [\mu^{2}_{i}(x)] -  \mathbb{E}_{i} [\mu_{i}(x)]^{2}\\
     &= \underbrace{\mathbb{E}_{i} [\sigma^{2}_{i} (x)]}_{Aleatroric 
     \,\,uncertainty} &+ \underbrace{\mathrm{Var}_{i}[\mu_{i}(x)]}_{Epistemic \,\, uncertainty}
 \end{aligned}
 \label{uncertainty_disentanglement}
\end{equation}
Equation \eqref{uncertainty_disentanglement} shows that across multiple output samples, the mean of variances represents aleatoric uncertainty, while the variance of mean represents the epistemic uncertainty. The predictive variance, $\sigma_{i}^{2}(x)$ is obtained using the Gaussian NLL loss \eqref{beta_NLL}. However, predictive uncertainty only accounts for the data and model uncertainty during future trajectory prediction and does not have the information of upstream perceptual  uncertainty obtained using KF. In order to capture the perceptual uncertainty, the NN is trained on augmented trajectory samples that takes both state and associated covariance as inputs.   The perceptual uncertainty is then estimated by minimising the MSE loss  between actual covariance obtained using KF with the predicted covariance from NN. Details of the method and results  have  been shown in section \ref{state_unc}.


\renewcommand{\thealgorithm}{}

  \begin{algorithm}
   \caption{}
    \begin{algorithmic}
      \Function{Kalman}{$\mathbf{X_{k-1}},\mathbf{P_{k-1}},\mathbf{R},\mathbf{Q}$}\Comment{Where $\mathbf{X_{k-1}}$ - state, $\mathbf{P_{k-1}}$ - cov, $\mathbf{R}$ - measurement noise , $\mathbf{Q}$ - process noise}

        \For{$k = 1$ to $N$}
        
            \State $x_{k} = F  x_{k-1}$ \Comment{Predict Step}
            \State $P_{k} = F  P_{k-1}  F^{T} + Q$\\

            \State $S = H P_{k-1} H^{T} + R$
            
            \State $K = P_{k-1} H^{T} S^{-1}$
            \State $y_{k} =z_{k} -  H x_{k}$
            \Comment{Innovation}\\
            
            \State $x_{k} = x_{k} +  K y_{k}$ 
            \State $P_{k} = P_{k} - K H P_{k}$ \Comment{Update Step}\\

        \EndFor
        \EndFunction\\
    \For{$k = 1$ to $M$}\Comment{M samples for M ensembles}
                    \Function{Trajectory Sampling}{$\mathbf{X_{k}},\mathbf{P_{k}}$}
                    \State$x_{sample} = \mathrm{Multivariate Normal}(\mathbf{X_{k}},\mathbf{P_{k}}$)
             \EndFunction
    \EndFor\\

    \Function{Model}{input = $\mathbf{[X_{k}, \Sigma_{k}]^{T}}$, target = $\mathbf{[y_{k}, \Sigma_{k}^{y}]^{T}}$, num\, epochs, batch,  M}\Comment{End-to-End Training Model}
        \For{$epoch = 1$ to $num\,epochs$}
        \State $\mathbf{[\hat{y}_{k}, \hat{\Sigma}_{k}^{s}, \hat{\Sigma}_{k}^{p}]} = \mathrm{Model}(\mathbf{[X_{k}, \Sigma_{k}]^{T}})$\Comment{Outputs}
        \\
        \State$MSE = \lVert \mathbf{\hat{\Sigma}_{k}^{s} - \Sigma_{k}^{y}} \rVert$ 
        \State$NLL = \dfrac{\lVert \mathbf{y_{k} - \hat{y}_{k}} \rVert}{\mathbf{\hat{\Sigma}_{k}^{p}}} + \dfrac{log(\mathbf{\hat{\Sigma}_{k}^{p}})}{2} $ 
        \EndFor
    \EndFunction
\end{algorithmic}
\end{algorithm}

\section{Experiments}
In this section, we discuss  the datasets, data augmentation,  implementation details for each network and the performance metrics.  Following common practice from literature \cite{Alahi}, we trained our models on publicly available pedestrian datasets. Two most popular datasets are the ETH dataset \cite{Pellegrini} which contains the ETH and HOTEL scene while the UCY dataset \cite{UCY} which contains the UNIV, ZARA1 and ZARA2 scenes.    In order to draw parallelism with past works \cite{Nikhil}, we studied 8 (3.2 secs) historical steps to predict 12 (4.8 secs) steps into the future. 

\subsection{Data Augmentation}
Initially, we trained our model  on the ETH  dataset only which contains approximately 420 pedestrian trajectories under varied crowd settings. However, a small number of trajectories is insufficient for training. Therefore, we performed  data augmentation using Taken's Embedding theorem \cite{Takens}. We used a sliding window of T = 1 step to generate multiple small trajectories out of a single large trajectory. For instance, a  pedestrian's  trajectory of 29 steps will result in  10 small $\{x,y\}$ trajectory pairs of 20 steps each if past trajectory information of  8 steps is used for predicting  12 steps into future. In total, we constructed 1597 multivariate time series sequences which we split into 1260 training  and 337 testing sequences for the ETH hotel dataset. Further, each trajectory was augmented using KF to generate posterior state and covariance distribution. Then, M trajectories were sampled from the distribution for each original trajectory. Details of data augmentation using KF and TS have been discussed in sec.\ref{sec:methods}.

\subsection{Implementation details}
The encoder-decoder neural network was trained end-to-end using PyTorch. Adam optimizer with  a learning rate of $1e-3$ was used  to compute the  MSE and  NLL loss.  MSE loss was minimized to estimate the covariance of KF while NLL loss was minimized to capture the predictive uncertainty. Each model was trained for 150 epochs with a batch size of 64. For the ensemble model, M=3 networks were considered while for the MC dropout, a single model with dropout probability, p = 0.5 was used based on our previous research \cite{Nayak}.   The model was compiled and fit using train data and test data was used for predictions.  

\subsection{Performance Metrics} \label{perf_m}
The trained model is then used to predict the distribution for pedestrian future states. 
Overall, the predictions are averaged to generate the mean predicted path along with the associated  variance that quantifies uncertainty.  We adopt the widely used performance metrics \cite{Alahi} namely average displacement error (ADE) and final displacement error (FDE)  for prediction comparison between the ground truth  and mean predicted path. Further, we define valid prediction intervals for regression problems based on performance metrics like prediction interval coverage probability (PICP) and mean prediction interval width (MPIW) \cite{Dewolf}.  \hfill
\vspace{0.1cm}

(a) Prediction Interval Coverage Probability (PICP): Coverage probability for a single state shows  whether the ground truth state, $\mathbf{y_{k}}$ lies within the predicted covariance ellipse  $\Gamma(X_{k})$ for the state $\mathbf{X_{k}}$, 
\begin{align}
    \mathcal{C}(\Gamma)\approx \frac{1}{|\mathcal{D^{*}}|} \sum_{(x,y \in \mathcal{D^{*}})}^{} \mathbbm{1}(y_{k} \in \Gamma(X_{k}))
    \label{eqn:PICP}
\end{align}
$\mathbbm{1}$ denotes an indicator function representing  Boolean values.

(b) Mean Prediction Interval Width (MPIW): It refers to the average width of the confidence interval. For the current results, we consider MPIW as  the average of the major and minor axes of the predicted covariance ellipse. 

\begin{align}
    \mathcal{W}(\Gamma)\approx \frac{1}{|\mathcal{D^{*}}|} \sum_{(x,y \in \mathcal{D^{*}})}^{} {(|u(x) - l(x)|)}
    \label{eqn:MPIW}
\end{align}
$u(x)$  and $l(x)$ refer to the lower and upper bounds for the prediction interval.

(c) Average Displacement Error (ADE):  Mean  Euclidean distance between predicted and ground truth. 

\begin{align}\label{ADE}
\begin{split}
    \text{ADE} &= \frac{1}{T}\sum_{t=t_{0}}^{t_{f}}
    ||{\mathbf{\hat{y}}_{(t)}- {\mathbf{y}}_{(t)}}|| 
    \end{split}
\end{align}

(d) Final Displacement Error (FDE):  Euclidean distance between the predicted  and true final state across all trajectories.

\begin{align}\label{FDE}
\begin{split}
    \text{FDE} &= 
    ||{\mathbf{\hat{y}}_{(t_{f})}- {\mathbf{y}}_{(t_{f})}}|| 
\end{split}
\end{align}

where $\mathbf{\hat{y}}_{t}$ is the predicted location at timestamp t and $\mathbf{y}_{t}$ is the ground truth position. 

\section{Results}

\subsection{Why Ensemble?}
For quantifying uncertainty,  deep ensembles  average predictions over an ensemble of independently trained networks. In the current scenario, each network is trained using the Gaussian negative log-likelihood (NLL) \eqref{beta_NLL} loss function such that the network outputs probabilistic predictions with both mean ($\mu$) and variance ($\sigma^{2}$). As, a single network can generate probabilistic outputs when trained with NLL loss function,   \textbf{ why consider averaging the predictions over an ensemble of M networks?} To answer this question, we observe how the  NLL and test MSE loss  scale with the  number of independent networks (M)  within an ensemble. The losses were evaluated on the ETH \cite{Pellegrini} dataset for pedestrians.   
\begin{table}[h!]
  \begin{center}
    \caption{Scalability of NLL (nats) and MSE with number of networks (M) within an ensemble}
    \label{tab:table2}
    \begin{tabular}{c|c|c} 
      \textbf{M} & \textbf{NLL} & \textbf{MSE}\\
      \hline\\
      1 & -0.335 & 0.214\\
      2 & -0.362 & 0.205\\
      3 & -0.377 & 0.205\\
      4 & -0.371 & 0.208\\
      5 & -0.379 & 0.200\\
    \end{tabular}
  \end{center}
\end{table}

Each neural network within the ensemble was randomly initialized at the beginning of the training. Additionally, for each network, a training trajectory was randomly sampled as input from the distribution of trajectories. Training was performed  and the final NLL loss  was averaged over the number of networks, M within the ensemble. Test MSE was evaluated on a set of test trajectories different from the training data after the model was fully trained   (Table \ref{tab:table2}). The results indicate an ensemble of five networks had the lowest NLL as well as test MSE. Indeed, this shows an ensemble network because of  lower NLL captures better predictive uncertainty. Further, low test MSE shows predictions of an ensemble network are closer to ground truth as compared to a single network. Overall,  any ensemble of networks with $M>2$ produced better NLL and MSE as compared to a  single network. 

\subsection{Predictions: single vs Ensemble} \label{no_input_unc}
\begin{figure}[h!]

 \centering
\begin{minipage}{0.45\textwidth}
   \includegraphics[width=1\linewidth]{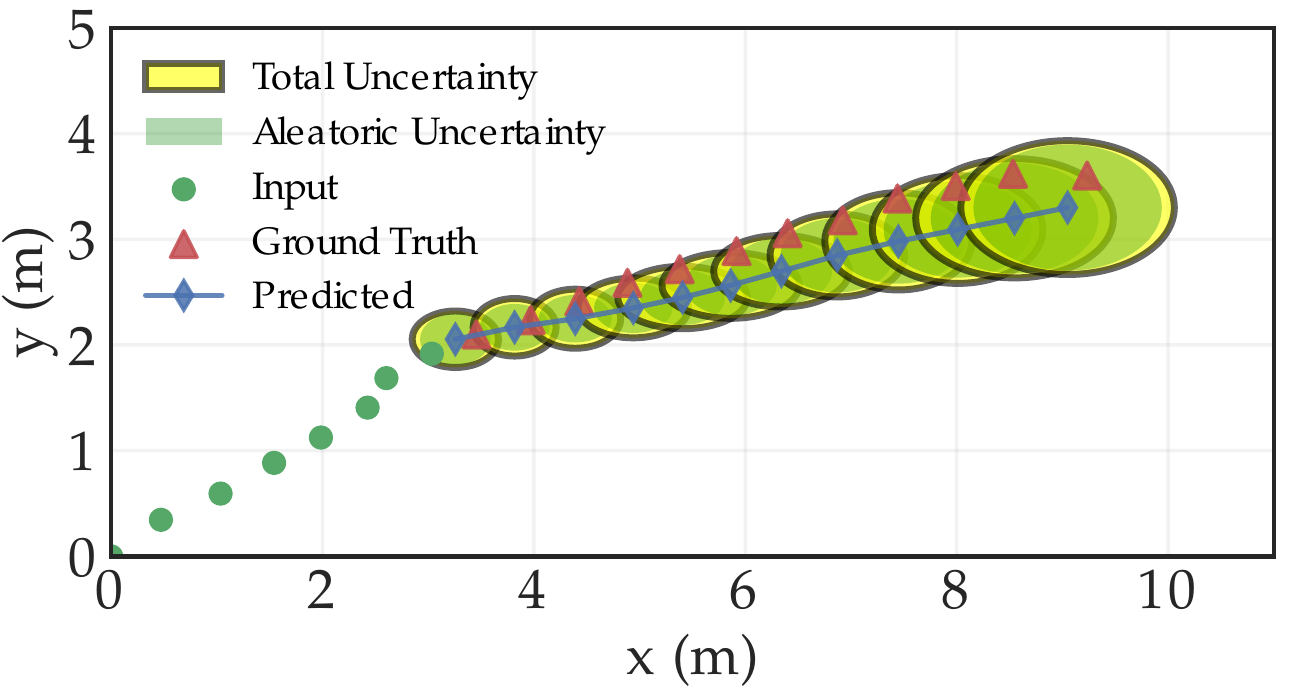}
   \label{fig:3b} 
\end{minipage}
\vfill
\begin{minipage}{0.45\textwidth}
   \includegraphics[width=1\linewidth]{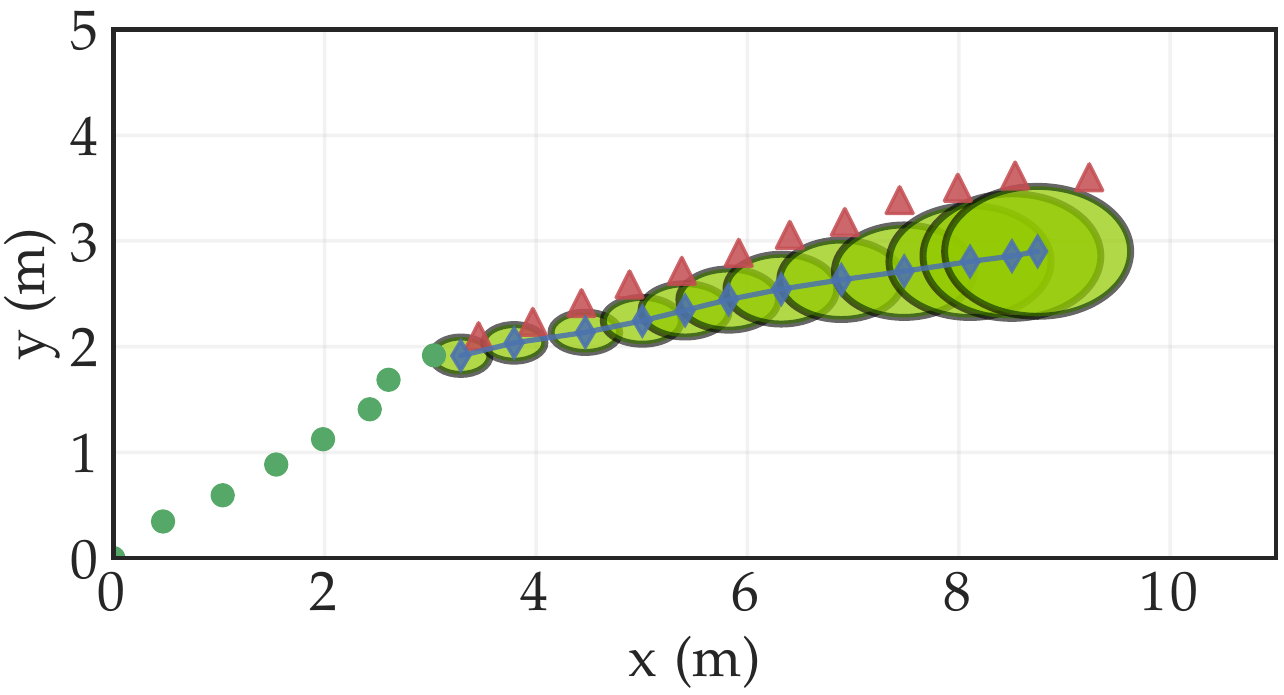}
   \label{fig:3a}
\end{minipage}

        



  \caption{Predictive uncertainty for (a) Deep Ensemble with M=3 networks (b)  single network.}\label{fig:3}
\end{figure}

In this section, we compare the predictive uncertainty of a single network with an ensemble of three networks (M=3) on a pedestrian trajectory from the ETH dataset.   Figure \ref{fig:3} shows the predictive uncertainty. The model takes 8 input states ({\color{green} \transparent{0.5}{$\bullet$}}, green dot) to predict 12 states into future.  {\color{red} \transparent{0.65}$\blacktriangle$}  represents the actual ground truth trajectory of the pedestrian. Further, the plot shows the mean predicted path ({\color{blue} \transparent{0.75}{$\blacklozenge$}}, blue diamond)  alongwith the  $1\sigma$  covariance ellipse 
 to quantify uncertainty during prediction.

Figure \ref{fig:3}b shows the predictive uncertainty for a single network, which fails to generate accurate prediction interval with respect to the ground truth. A significant portion of the ground truth trajectory lies outside of the $1\sigma$ predictive covariance. On the other hand, the ensemble network (Figure \ref{fig:3}a) produced better predictive uncertainty and mean path by averaging the mean and variance of predictions over an ensemble of networks. The plot shows that the ground truth completely lies within the confidence interval at each time step. Further, the ADE/FDE for the ensemble network (0.618/1.137) was significantly lower compared to the ADE/FDE for a single network (0.704/1.394).


Figure \ref{fig:3}a shows the total predictive uncertainty, which is due to the combination of aleatoric and epistemic uncertainty. The aleatoric uncertainty represents the inherent noise in the data, while the epistemic uncertainty arises due to the variation in model predictions. Since, the test data is sampled from the same distribution as the train data, the model uncertainty highlighted in yellow, is  negligible compared to the aleatoric uncertainty.   In contrast, a single network has no model uncertainty, and the total predictive uncertainty and aleatoric uncertainty are the same. Thus, the ensemble network is better suited to handle epistemic uncertainty, which is critical for robust  real-world applications.


\begin{figure}[ht]
   \centering      
   \includegraphics[width=0.4\textwidth]{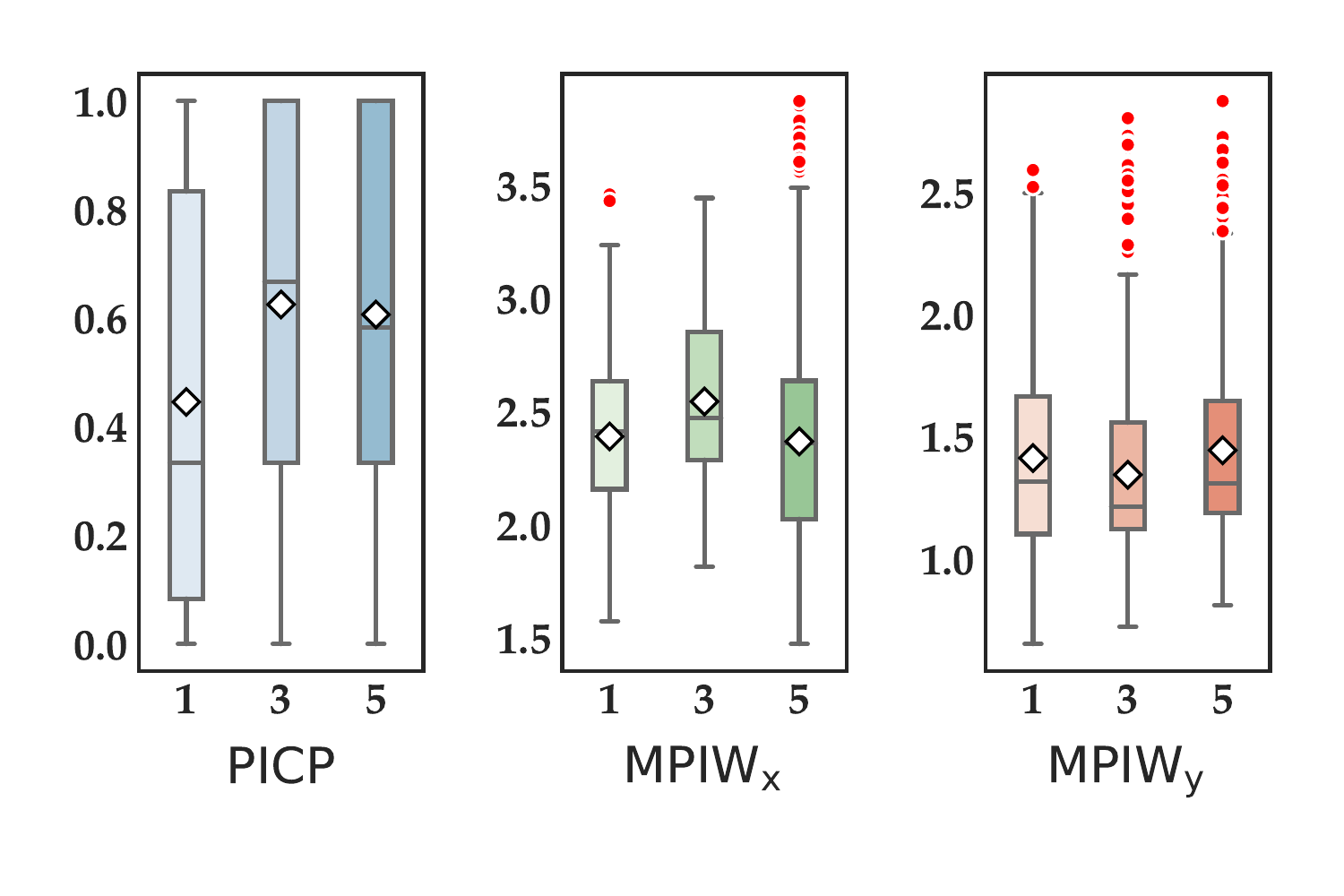} 
 \caption{Variation of prediction metrics (a) PICP (b) $\mathrm{MPIW_{x}}$ (c) $\mathrm{MPIW_{y}}$ with ensemble models.}
 \label{fig:Fig4}
 \end{figure}

Further, we compare the performance metrics, coverage probability  PICP \eqref{eqn:PICP} and prediction interval width  MPIW   \eqref{eqn:MPIW} for different ensemble networks with the ETH dataset (Figure \ref{fig:Fig4}). 
 Results show that the ensemble network with M = $\{3,5\}$ networks provide better predictive uncertainty estimates than a single network, with an average coverage probability of $ \approx 63\% $ compared to $45\%$. In addition, current results also reveal that the MPIW for an ensemble model with multiple networks is either comparable or less than that of a single network, indicating that even with a smaller confidence interval, the ensemble model can achieve a higher coverage probability for the predictions. We denote the average width of major and minor axes as $\mathrm{MPIW_{x}}$ and $\mathrm{MPIW_{y}}$ respectively.
 
\subsection{Incorporating Perception uncertainty}\label{state_unc}


Previous studies have focused on capturing predictive uncertainty while neglecting upstream state uncertainty during perception. Incorporating perception or state uncertainty into the prediction pipeline remains a challenge, as it is unclear how the total predictive uncertainty will be affected. To address this challenge, we propose incorporating and propagating state uncertainty by including the associated state covariance, $\bar{P_{k}}$, obtained at each step from the KF. We append the heteroskedastic noise associated with each state to the state, $\mathbf{X_{k}}$, and train them together. In  section \ref{no_input_unc}, the state $\mathbf{X_{k}}$ =  $[x,y]$ contained only respective states sampled from the posterior distribution of state covariance using KF. Here, we have neglected the velocity, [u,v] in the states  for training as no significant improvement was observed with their inclusion.  In the current scenario, we append the states and the associated covariance together as $\mathbf{[X_{k}, \Sigma_{k}]^{T}}$ and train them jointly.  The outputs are $\mathbf{[\hat{y}_{k}, \hat{\Sigma}_{k}^{s}, \hat{\Sigma}_{k}^{p}]^{T}}$. $\mathbf{\hat{\Sigma}_{k}^{s}}$ corresponds to the estimated state covariance which is trained by minimizing the MSE loss with the actual covariance, $\mathbf{\Sigma_{k}}$ obtained using KF. This enables the NN  capture the upstream perceptual uncertainty. Meanwhile, $\mathbf{\hat{\Sigma}_{k}^{p}}$  corresponds to the estimated predictive covariance by the NN and was obtained by minimising the NLL loss for the state $\mathbf{X_{k}}$.  Overall, our method enables us to estimate the state uncertainty and incorporate it into the prediction pipeline, which can help improve the total  uncertainty estimation.

\subsubsection{Perception uncertainty}

Sensing uncertainty during  state estimation of a dynamic object  can be obtained recursively from  sensor measurements using  KF. 
However, estimating perceptual uncertainty for future  states over a long prediction horizon using  KF can be challenging. Further, the uncertainty represented by covariance will continuously grow based on the motion model.  To address these issues, a neural network (NN) model is trained to learn the KF for estimating perceptual uncertainty at any future state. We perform domain randomization on train trajectories by varying the measurement covariance, $\mathbf{R} \in (2\%, 20\%)$ to  generate  trajectories across a wide range of sensor noise. It will make the NN more robust in estimating  the sensing uncertainty effectively.  

\begin{figure}[ht]
   \centering      
\includegraphics[width=0.4\textwidth]{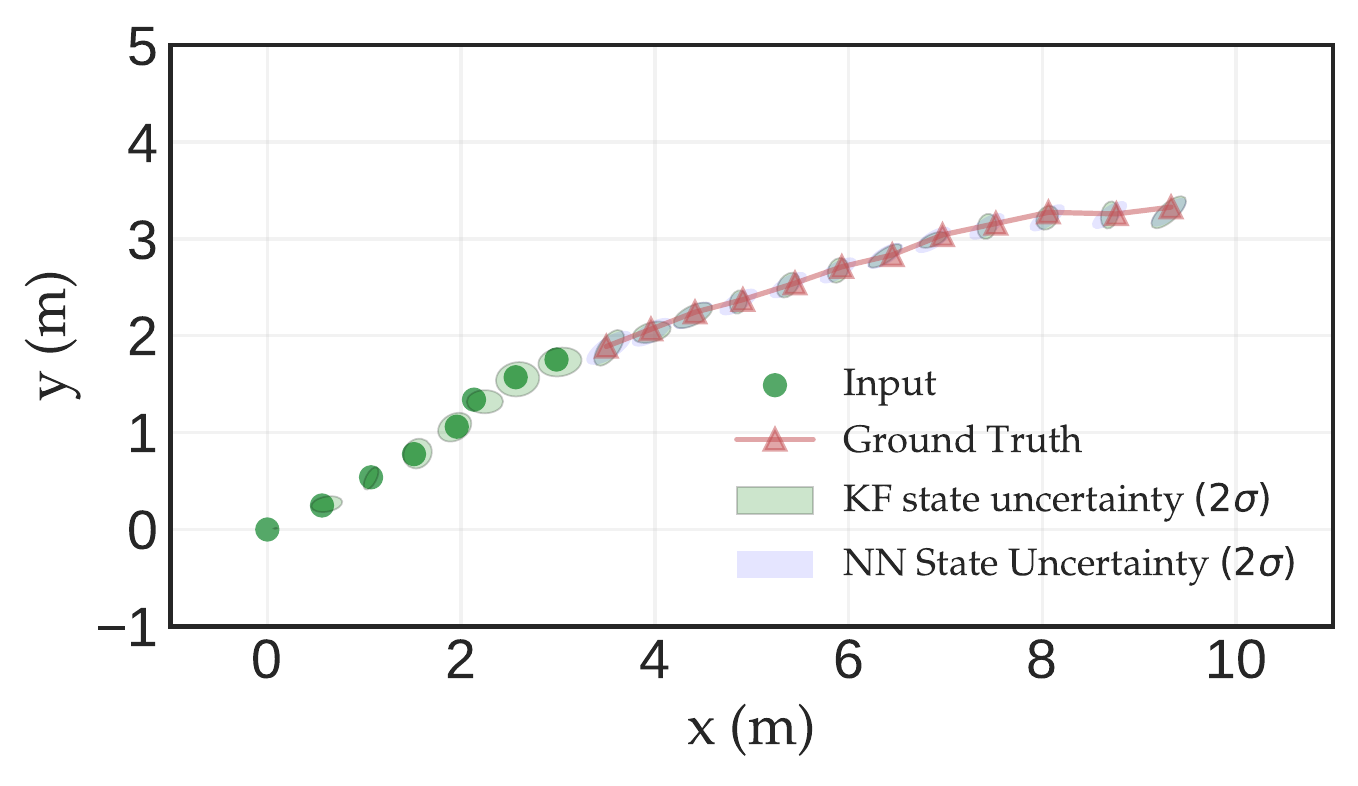} 
 \caption{ Covariance estimation using neural network capturing sensing uncertainty. }
 \label{fig:Perceptionunc_secII}
 \end{figure}

 To achieve this, we apply  KF  to both the input and ground truth  of every single trajectory to obtain the estimated state and covariance. The resulting states and covariance are then concatenated and trained together using an encoder-decoder network. The network minimizes the mean squared loss between KF ground truth covariance, $\mathbf{\Sigma_{k}}$  and predicted covariance, $\mathbf{\hat{\Sigma}_{k}^{s}}$. Figure \ref{fig:Perceptionunc_secII} depicts the outputs of the neural network for $\mathbf{R} = 5\%$ that closely resembles the covariance predictions of the KF.  The NN takes 8 states with associated covariance obtained using the KF as input and predicts the 12 future states, as well as the estimated state covariance due to sensing uncertainty. Our results demonstrate that the $2\sigma$ confidence interval predicted using the neural network closely matches the state uncertainty on ground truth obtained using the KF. This capability allows the NN to estimate the state covariance associated with perception uncertainty at each future state, which can then be integrated with the prediction uncertainty to enhance the system's robustness.

\subsubsection{Prediction uncertainty}

\begin{figure}[h!]
   \centering      
\includegraphics[width=0.4\textwidth]{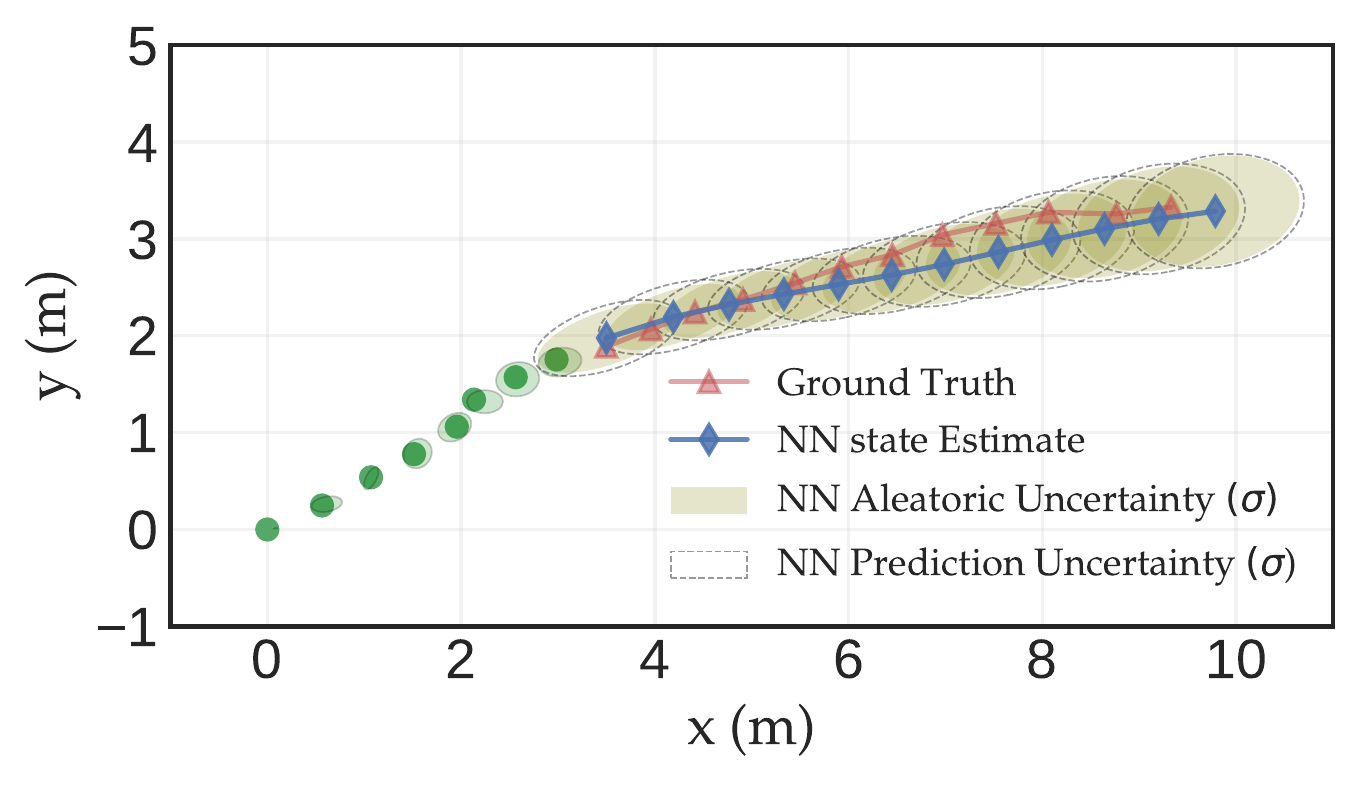} 
 \caption{Total Predictive uncertainty}
 \label{fig:Prediction_unc_secII}
 \end{figure}

Unlike perception uncertainty which accounts for noise in the process or measurement during sensing, prediction uncertainty captures the unpredictability  associated with future states. In Figure \ref{fig:Prediction_unc_secII},  we show the total predictive uncertainty for the same trajectory as before. For capturing predictive uncertainty, we train the network using NLL loss and the NN outputs both the mean, $\mu_{k}$ and covariance, $\mathbf{\hat{\Sigma}_{k}^{p}}$ for the predicted distribution.  We treat $\mathbf{\hat{\Sigma}_{k}^{p}} = [\hat{\Sigma}_{xx}, \hat{\Sigma}_{xy}, \hat{\Sigma}_{yx}, \hat{\Sigma}_{yy}]$ as the full state covariance of a bivariate distribution.

We generate results for an ensemble of 3 networks and average the predicted distributions to obtain the mean predicted path and uncertainty at each state. The average ADE/FDE of the mean predicted path across all test trajectories is 0.64/1.08.  
 For the ensemble network, the mean of variance,$\mathbf{\hat{\Sigma}_{k}^{p}}$, represents aleatoric uncertainty across the ensemble. Meanwhile, the variance of predicted means represents the "model" or epistemic uncertainty. As the test samples are from the same distribution as train samples, the variation in model predictions is insignificant, and thus the epistemic uncertainty is negligible too. Further, the predictive uncertainty around each state is significantly larger than the predicted sensing uncertainty.

\subsubsection{State and Prediction Uncertainty}

\textbf{Why incorporate sensing uncertainty into prediction pipeline?}

The primary objective of this paper is to design an end-to-end estimator that can effectively estimate state uncertainty by taking  noisy sensor measurements and propagate the state uncertainty into the future predicted states. This approach enables the neural network to make precise future state  predictions while remaining robust to upstream uncertainty. Mathematically, we formulate total uncertainty as the combination of sensing and predictive uncertainty.

\begin{figure}[h!]
\includegraphics[clip, trim=0.25cm 2cm 0.25cm 4cm,width=0.45\textwidth]{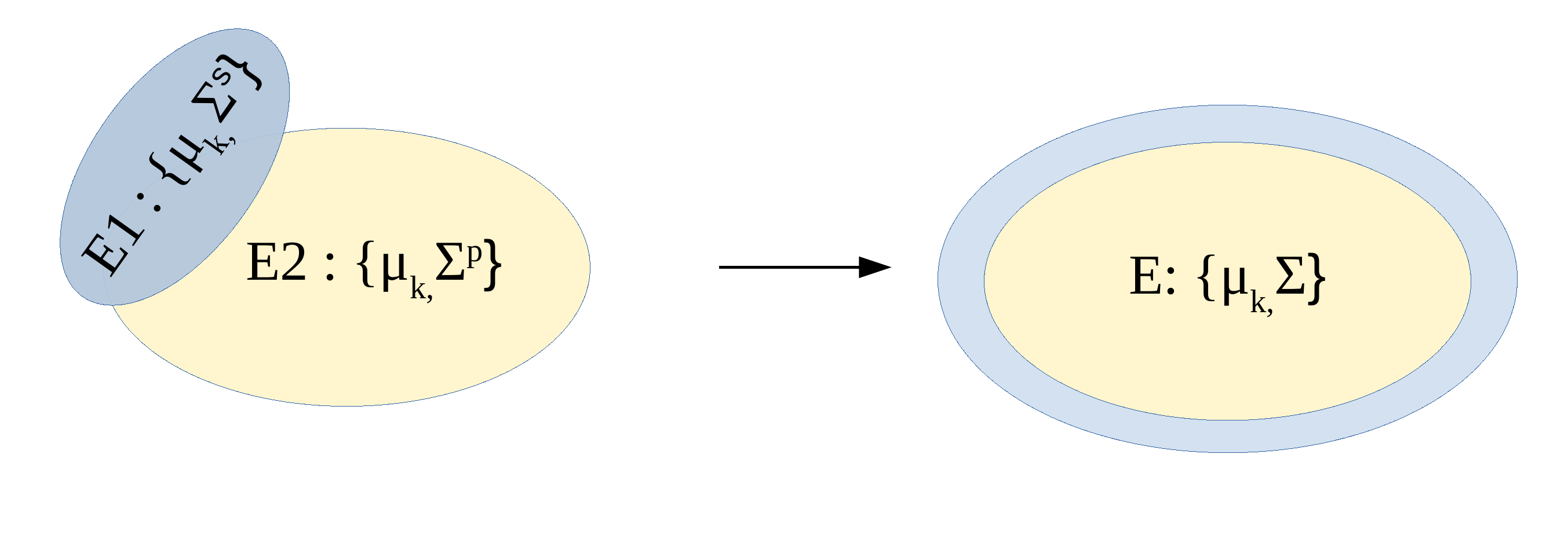}
\caption{Total uncertainty  corresponds to the Minkowski addition of prediction and state uncertainty.}
\end{figure}

Assume, the upstream state uncertainty  due to noisy measurements is represented by the covariance ellipse, E1. 
\begin{equation*}
    E1 = \{ x_{1}{ \in \mathbb{R}^{2}} : \, ( x_{1} - \mu_{k})^{T}(\mathbf{\hat{\Sigma}_{k}^{s}})^{-1} ( x_{1} - \mu_{k}) \leq 1\}
\end{equation*}
Similarly, at each time,  a state randomly sampled from the  covariance ellipse representing prediction uncertainty as:
\begin{equation*}
     E2 = \{ x_{2}{ \in \mathbb{R}^{2}} : \, ( x_{2} - \mu_{k})^{T}(\mathbf{\hat{\Sigma}_{k}^{p}})^{-1} ( x_{2} - \mu_{k}) \leq 1\}
\end{equation*}

Covariance ellipses E1 and E2 represent  convex polytope of all reachable states during perception and prediction respectively.  Any sampled state, $x_{2}$ from the predictive uncertainty, $\mathrm{E2} = \mathbf{\hat{\Sigma}_{k}^{p}}$ represents a possible  future state of the tracked object. However, the sampled state has no information of upstream uncertainty, $\mathrm{E1} =\mathbf{\hat{\Sigma}_{k}^{s}}$. Therefore, perceptual uncertainty can be incorporated into the prediction uncertainty 
 as the Minkowski addition of the closed convex polytopes, $\mathrm{E1}$ and $\mathrm{E2}$ centred around the origin \eqref{eq:tot_unc}.

\begin{equation}
E = \{ x_{1} \oplus x_{2} : \, x_{1} \in E1, \,x_{2} \in E2\}
\label{eq:tot_unc}
\end{equation}

Here $\oplus$ denotes the vector addition. Further, we translate the summed covariance ellipse representing total uncertainty to the mean predicted state, $\mu_{k}$.

\begin{equation*}
   E' = \{ x + \mu_{k} : \,x \in E\}
\end{equation*}

Overall, $\mathrm{E'}$ represents the total reachable set of states for the end-to-end estimator at any instance.







 \begin{figure}[ht]
   \centering      
\includegraphics[width=0.4\textwidth]{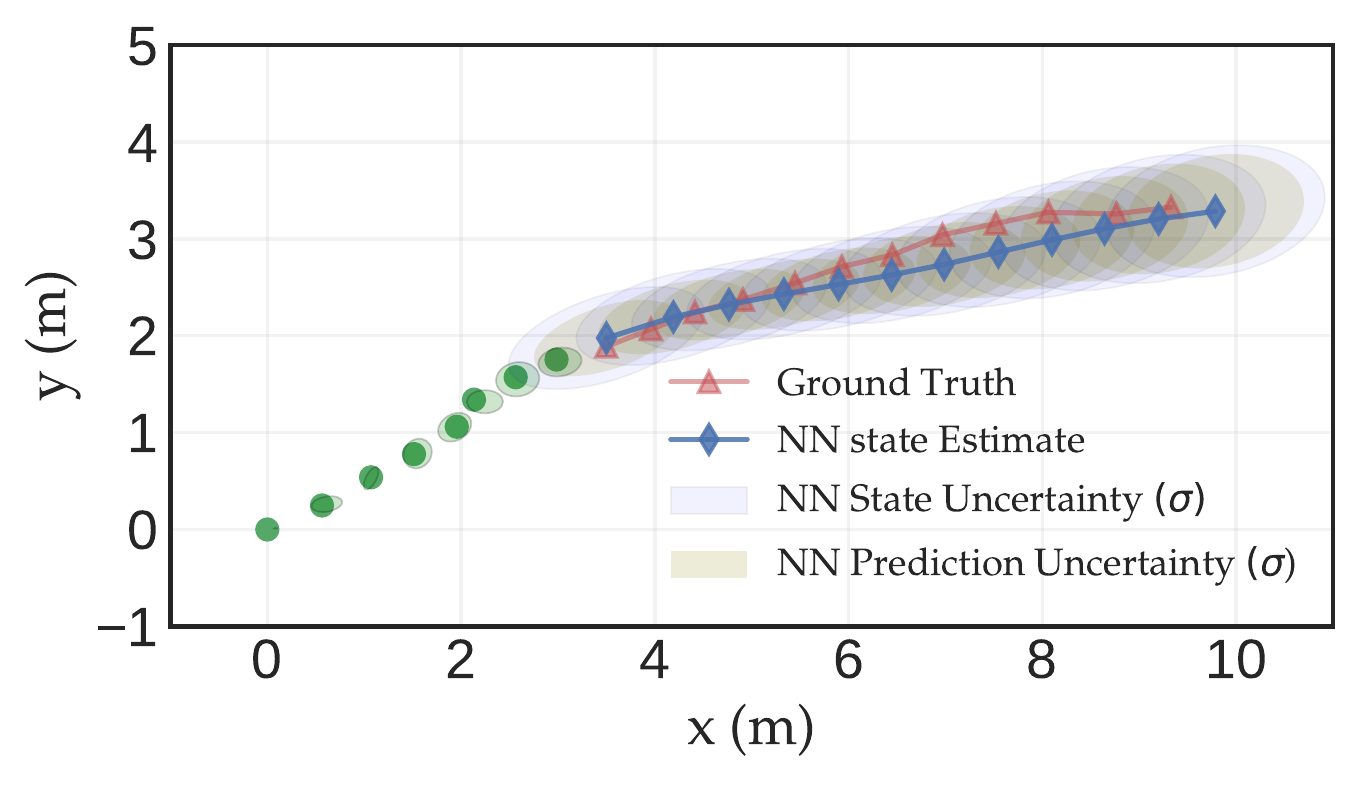} 
 \caption{Total uncertainty by incorporating state uncertainty into prediction}
 \label{fig:Total_unc_secII}
 \end{figure}

By including state uncertainty, the total uncertainty now becomes the Minkowski sum of state and prediction uncertainty, as shown in Figure \ref{fig:Total_unc_secII}. We evaluated the coverage probability for an ensemble of three networks, and with consideration of predictive uncertainty alone, the coverage probability was 0.63 (Figure \ref{fig:Fig4}). This predictive uncertainty was based on deterministic state inputs without considering any sensing uncertainty. However, when we accounted for measurement noise, $\mathrm{R} = 5\% $ for states and augmented the data using KF for training, the coverage probability improved by almost 20\% to 0.8. This demonstrates the effectiveness of our proposed approach in capturing both state and predictive uncertainty for accurate and robust future predictions.
\vspace{0.1cm}

\begin{figure*}[!htb]
 
        \minipage{0.3\textwidth}
          \includegraphics[width=\linewidth]{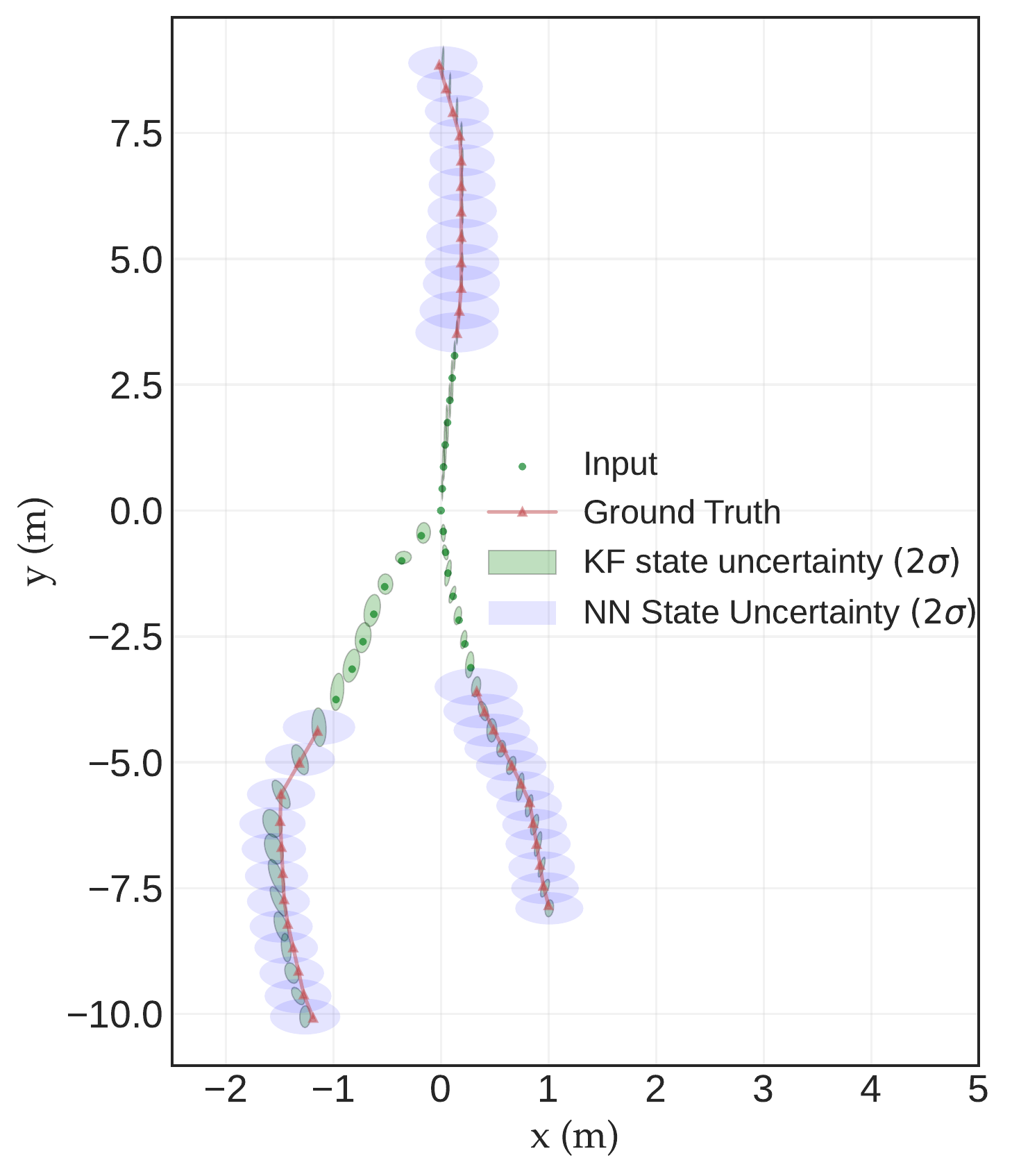}
        \endminipage\hfill
        \minipage{0.3\textwidth}
          \includegraphics[width=\linewidth]{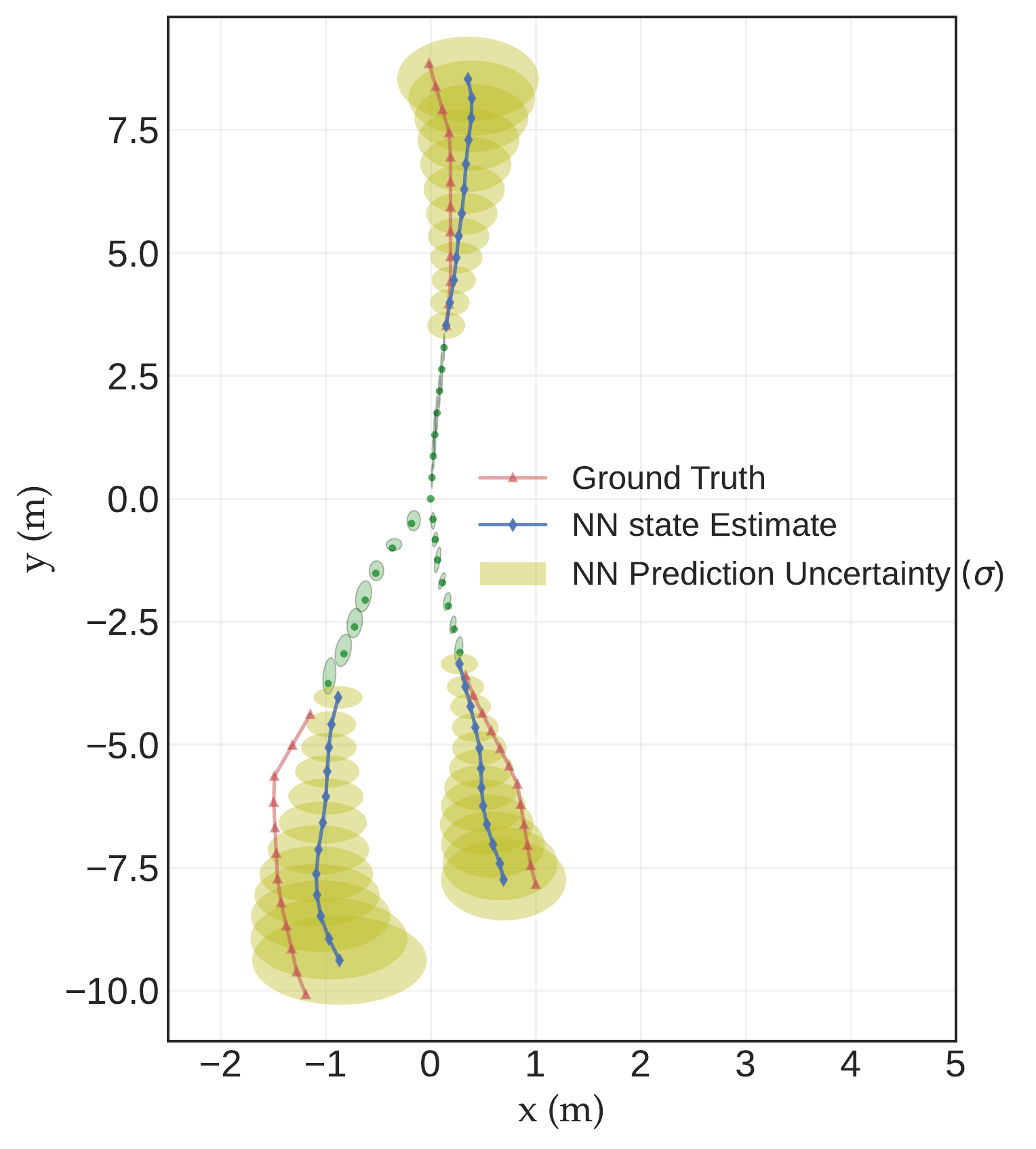}
        
        \endminipage\hfill
        \minipage{0.3\textwidth}%
          \includegraphics[width=\linewidth]{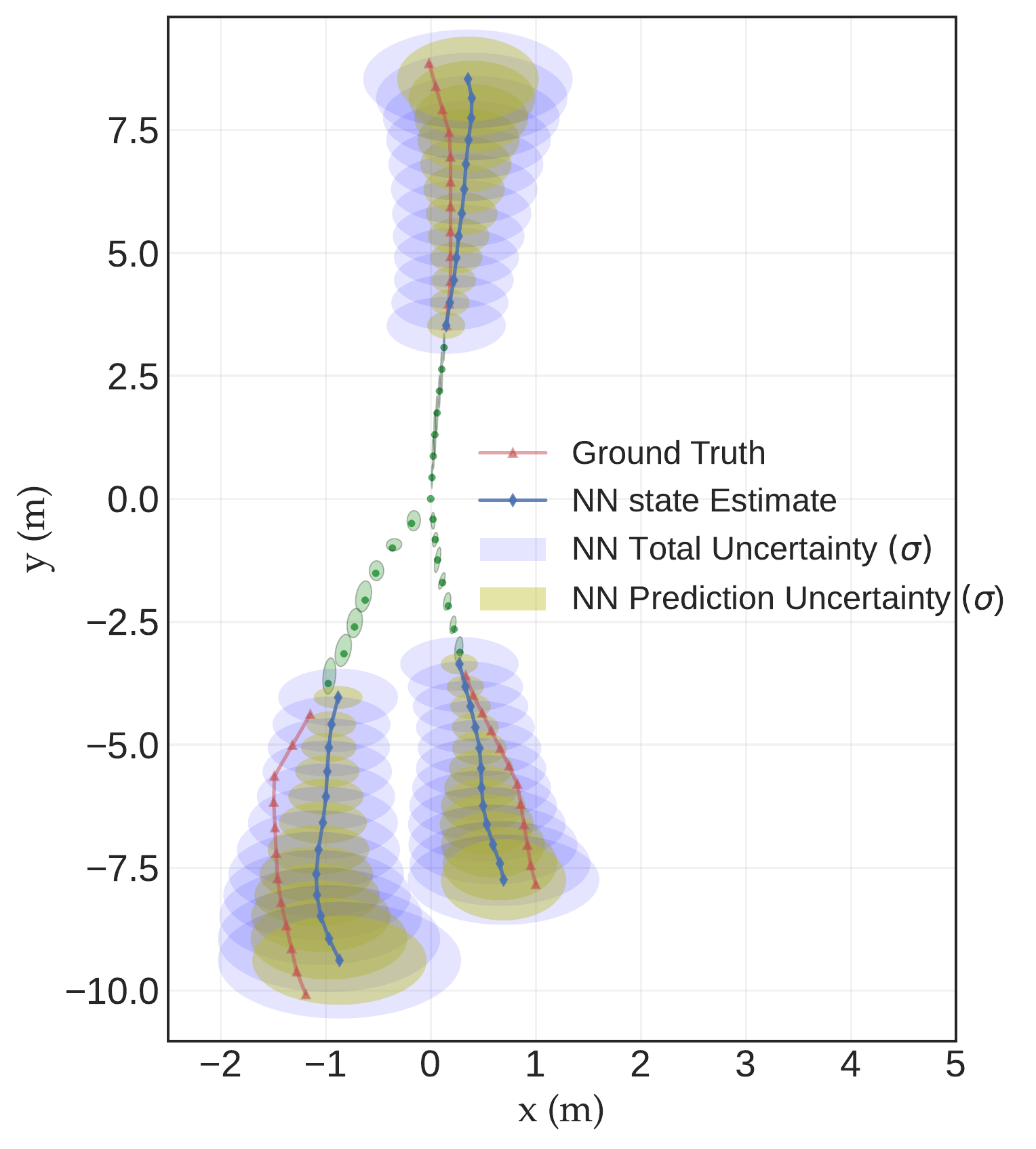}
          
        \endminipage

     \vfill
        \minipage{0.3\textwidth}
          \includegraphics[width=\linewidth]{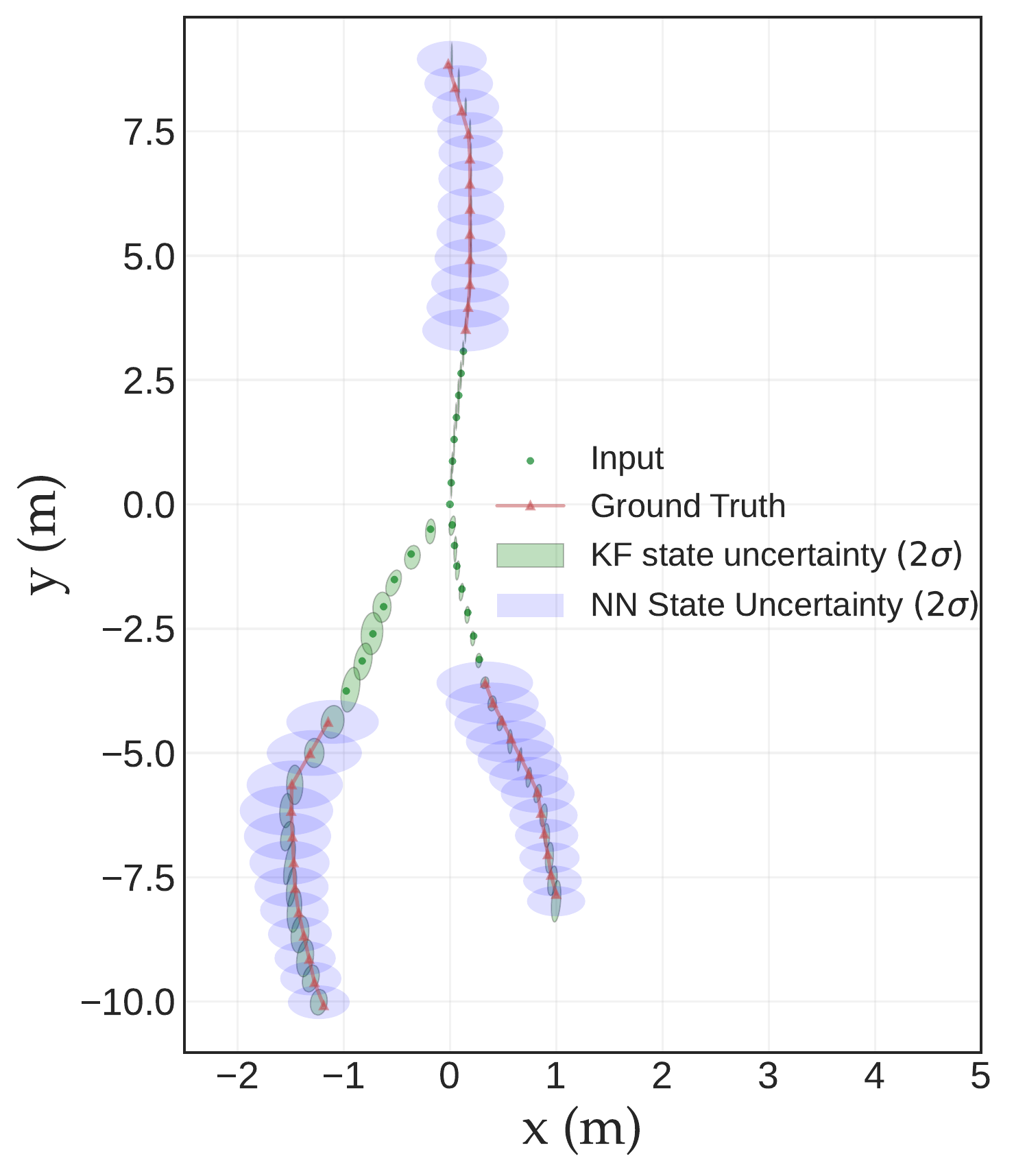}
        
        \endminipage\hfill
        \minipage{0.3\textwidth}
          \includegraphics[width=\linewidth]{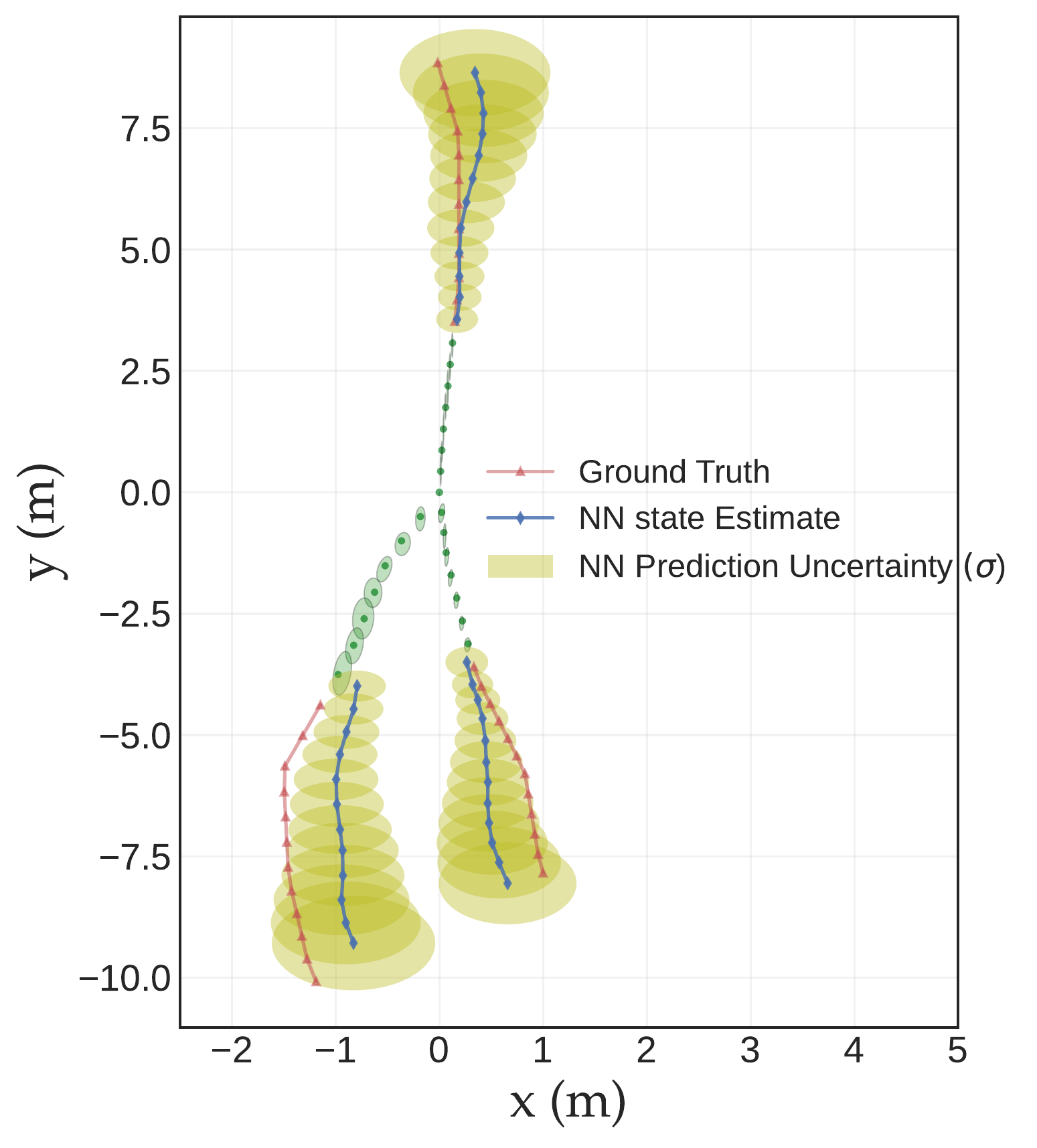}
          
        \endminipage\hfill
        \minipage{0.3\textwidth}%
          \includegraphics[width=\linewidth]{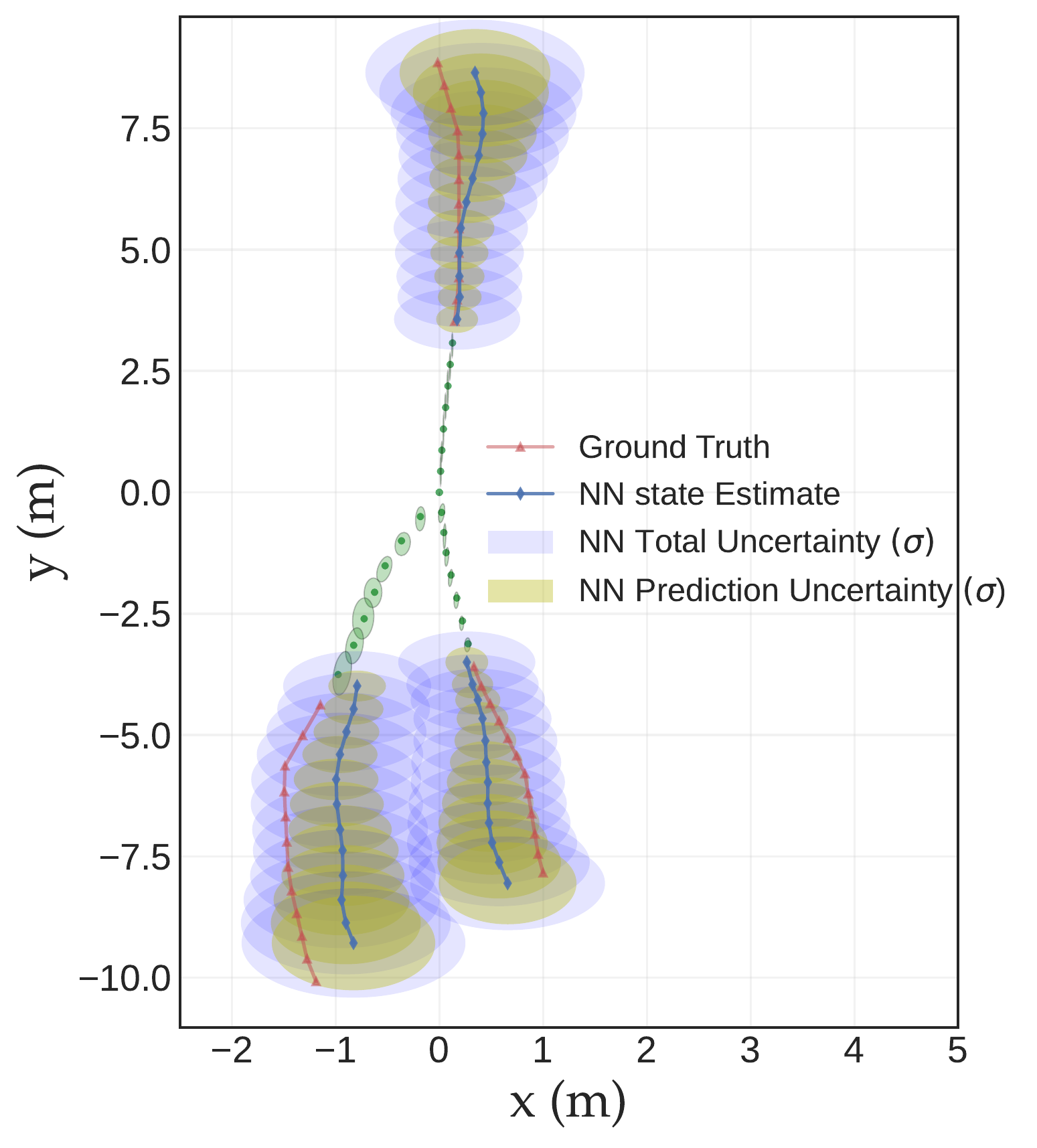}
        \endminipage

\caption{Comparison of $\mathbf{Left}$: Perception uncertainty, $\mathbf{Center}$: Prediction uncertainty, $\mathbf{Right}$: Total  uncertainty between $\mathbf{Top}$: Deep Ensembles  $\mathbf{Bottom}$: MC Dropout for trajectories from the ZARA 01 dataset.} \label{fig:Fig10}
\end{figure*}

\subsection{UQ Methods}

 In this section, we compare the prediction efficacy of Deep Ensemble with Monte Carlo (MC) dropout which is an approximate Bayesian inference method. The MC dropout leverages on the idea of training the NN  using  dropout layers and then performing inference at test time by randomly dropping  weights. This generates a distribution of  varying outputs instead of a single deterministic prediction. Like ensembles, the mean and variance of the output distribution can be computed to obtain the mean predicted path and quantify uncertainty. Details of the method has been described in  section \ref{sec:Dropout}.

We show the  the uncertainty plots for three test trajectories from the ZARA01 dataset, at $\mathbf{R}  = 5\%$ comparing both the methods (Figure \ref{fig:Fig10}). All trajectories start from origin. For deep ensemble, we consider M=3 networks while a dropout probability, p = 0.5 has been used on a single network for training using MC dropout method.

$\mathbf{Left:}$ The  plot compares the $2\sigma$   perception uncertainty between the NN and KF for each method. Both the methods are slightly under confident in predictions and overestimate the uncertainty bounds when compared with the KF ground truth. This may arise due to the simultaneous training using the NLL and MSE loss function. The NN fails to minimise the MSE loss function accurately while estimating covariance.  However, the  deep ensemble model slightly outperforms the  MC dropout in estimating the  perception uncertainty at each state. The MC dropout model overestimates the covariance associated with initial states. 

$\mathbf{Center:}$  The predictive uncertainty plot shows the $1\sigma$  distribution for future states with uncertainty bound for each method. The Predictive uncertainty can be disentangled into epistemic and aleatoric uncertainty. The uncertainty estimation is scalable and no significant difference was observed between the predictions of each  models. However, the  ADE and FDE for the mean predicted path of deep ensemble model (0.53/0.97) is closer to the ground truth when compared to the dropout model (0.58/1.00) as seen in Table \ref{tab:Table3}.  

$\mathbf{Right:}$ The top and bottom right plots show the $1\sigma$ total uncertainty for deep ensemble and MC dropout respectively. As discussed, the total uncertainty is the Minkowski addition of the covariance ellipses representing the perception and prediction uncertainty.  Since, the MC dropout overestimates the perception uncertainty, it affects the total uncertainty too. The MC dropout method makes under-confident predictions for total uncertainty. This phenomenon is more pronounced for the two trajectories along negative y-axis when compared to scalable predictions from deep ensembles.

Apart from uncertainty quantification, the current study also compared the performance metrics (sec. \ref{perf_m}) for both methods across the pedestrian datasets ETH\cite{Pellegrini} and UCY\cite{UCY}. The ADE/FDE results indicate that the deep ensembles have a closer mean predicted path to ground truth compared to MC dropout across all the datasets. Meanwhile, the $1\sigma$ coverage probability  results show deep ensembles have slightly higher coverage probability although not significant except for ETH dataset. 
 We have combined the prediction interval width across major, $\mathrm{MPIW_x}$ and minor, $\mathrm{MPIW_y}$ axes of predicted covariance ellipse to obtain the mean prediction interval width, $$MPIW = \sqrt{\frac{\mathrm{MPIW_x}^2}{2} + \frac{\mathrm{MPIW_y}^2}{2}}$$ Again, the deep ensembles have a lower MPIW compared to MC dropout which shows the deep ensembles are able to achieve slightly higher or equal coverage probability even with less prediction interval width. This shows that ensembles make robust predictions with scalable uncertainty and estimations closer to ground truth.

\begin{figure*}[ht]
   \centering      
\includegraphics[width=0.8\textwidth]{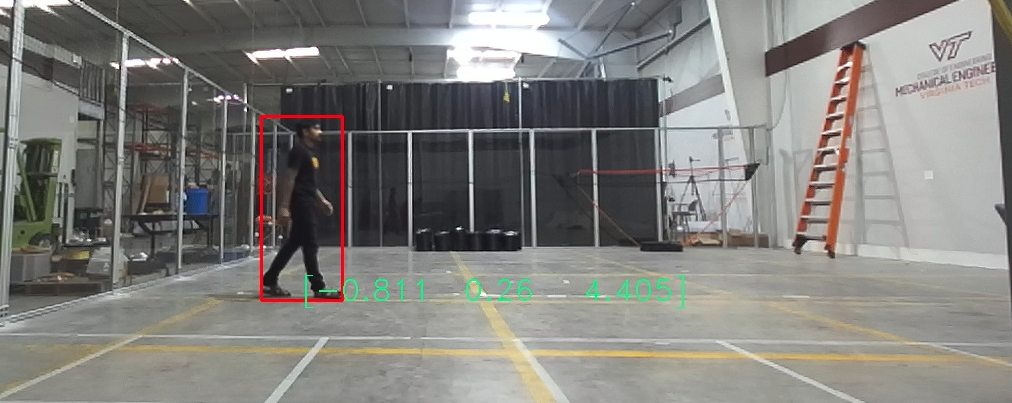} 
 \caption{Real world test trajectory set. Online tracking and estimation using a depth camera provides the position and velocity of the pedestrian in real-time. Generated test trajectory is used for offline prediction. }
 \label{fig:Real_traj}
 \end{figure*}

\begin{table}
\captionsetup[table]{skip=20pt}
\caption{Performance metrics showing ADE, FDE, PICP and MPIW for predicting 12 future time steps given 8 historical steps}
\setstretch{1.2}{\resizebox{0.45\textwidth}{!}{
\begin{tabular}{cc|c|c|c|c|}
&         & ADE  & FDE  & PICP & MPIW \\ \hline
\multicolumn{1}{c|}{{Deep Ensemble}} & ETH     & 0.60 & 1.11 & 0.80 & 2.23 \\
\multicolumn{1}{c|}{}                               & Hotel   & 0.40 & 0.67 & 0.88 & 1.44 \\
\multicolumn{1}{c|}{}                               & ZARA 01 & 0.53 & 0.97 & 0.81 & 1.82 \\
\multicolumn{1}{c|}{}                               & ZARA 02 & 0.56 & 1.12 & 0.87 & 2.08 \\
\multicolumn{1}{c|}{}                               & UNIV     & 0.25 & 0.50 & 0.93 & 1.55 \\ \hline
\multicolumn{1}{c|}{{Dropout}}       & ETH     & 0.7  & 1.2  & 0.73 & 2.38 \\
\multicolumn{1}{c|}{}                               & Hotel   & 0.44 & 0.66 & 0.88 & 1.46 \\
\multicolumn{1}{c|}{}                               & ZARA 01 & 0.58 & 1.00 & 0.8  & 1.83 \\
\multicolumn{1}{c|}{}                               & ZARA 02 & 0.59 & 1.15 & 0.86 & 2.06 \\
\multicolumn{1}{c|}{}                               & UNIV     & 0.27 & 0.54 & 0.94 & 1.59
\end{tabular}}

}

\label{tab:Table3}
\end{table}



\subsection{Out-of-distribution Results}

The current simulation results showed that the NN based end-to-end estimator yield  good performance for prediction on trajectories which follow same distribution as the training data. However,  one fundamental challenge for NN based prediction model  has been out-of-distribution (OOD) robustness.  Especially, \textit{if the test samples are based on real-world pedestrian trajectory with distributional shift, how effectively the trained NN model can predict the future state as well as the prediction and sensing uncertainty?} To test this hypothesis, we studied multiple scenarios namely walking fast, walking slow, turning left, turning right, walking normal  (Figure \ref{fig:Real_traj_pred}) which constitutes a  set of edge case scenarios which are different from the training samples.
\vspace{0.2cm}

\subsubsection{Test trajectory generation}
 In order to collect test  trajectories, we use the a depth camera recording at 30 frames per second (Figure \ref{fig:Real_traj}). The camera estimates the depth of the object based on a pair of images  to obtain the real-world position and velocity in 3D Cartesian coordinates. For object detection, we train a simple Mask R-CNN \cite{Kaiming} on the coco dataset \cite{Lin}. The object detection module accurately classifies the pedestrian and  tracks it real-time. The sampling time is set at 12 frames such that the camera obtains the object's position and velocity at every 0.4 seconds similar to  the simulation. Every single trajectory has a duration of 8 seconds resulting in 20 $\{x,y,u,v\}$ samples, out of which 8 samples(3.2 secs) represent past trajectory while 12 samples(4.8 secs) represent the ground truth.   We apply the current end-to-end estimator on the past trajectory to predict the future states with associated uncertainty and compare the predictions with ground truth.

\vspace{0.2cm}
\subsubsection{Sensor measurement noise}
    In order to estimate the measurement covariance, $\mathbf{R}$, we perform a simple calibration test. A single object was placed exactly 3m away from the camera. 60 samples pertaining to the $\{x,y\}$ position of the object were considered. The mean of the distribution was 2.9 $\pm$ 0.06 m. This shows roughly  $4\%$ noise on all measurements.   Kalman filter was applied to the each test trajectory for data augmentation based on the estimated measurement covariance for the sensor.

\vspace{0.2cm}
\subsubsection{Inference}
The weights of the NN model are trained on publicly available datasets namely ETH and UCY. Further, to ensure robustness towards varying degree of sensing noise, domain randomization was performed for measurement covariance, $\mathbf{R}$. As discussed previously, the trajectories are augmented using KF with a range of measurement covariance, $\mathbf{R} \in \{2-20\}\%$ and then trained using NN. This will make the NN model more robust towards prediction on test samples generated using a different sensor having a different   measurement noise.  During inference, only model parameters such as trained weights and biases were considered which makes the inference process computationally  cheap.  

\begin{figure*}
\centering
    \includegraphics[width=1\textwidth]{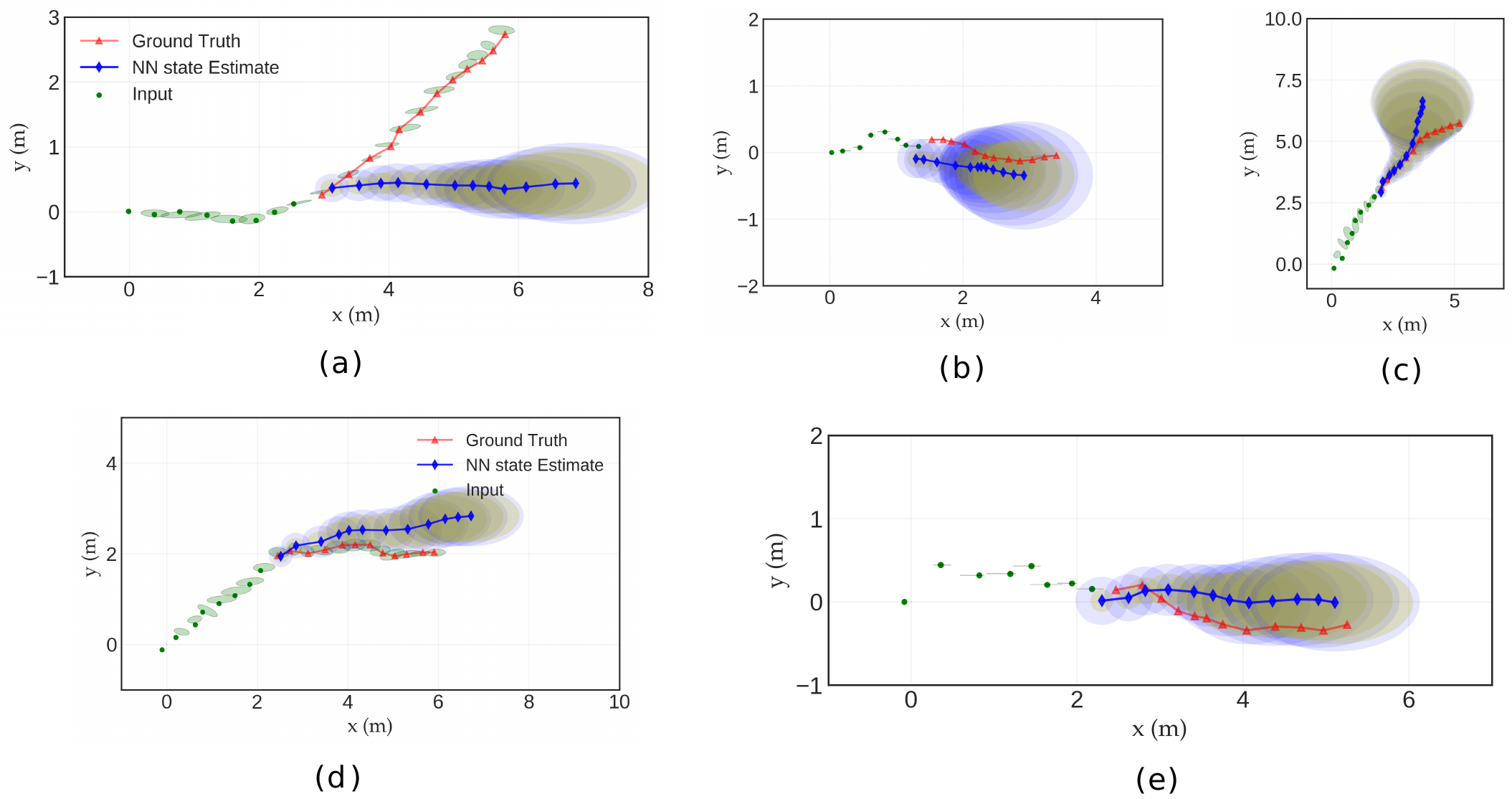} 
 \caption{Out-of-distribution prediction on multiple scenarios based on real-world pedestrian trajectory. The scenarios are : \textbf{(a)} Turning Left
\textbf{(b)} Walking Slow
\textbf{(c)} Walking Fast
\textbf{(d)} Turning Right
\textbf{(e)} Walking Normal. The covariance ellipse shows 1$\sigma$ total uncertainty disentangled into \textbf{blue}: perception uncertainty and \textbf{olive}: prediction uncertainty.}
\label{fig:Real_traj_pred}
\end{figure*}

Figure \ref{fig:Real_traj_pred} shows  end-to-end prediction on out-of-distribution trajectories for the considered scenarios. To draw parallel with simulation results, we predict 12 states into future based on 8 historical steps. The NN model based on deep ensembles with M =3 networks predicts both 1$\hat{\Sigma}$ sensing (blue) and prediction (olive) covariance ellipse alongwith the mean estimated path for each scenario. The ground truth trajectory lies within the predicted 1$\hat{\Sigma}$ total covariance ellipse except for the left turn trajectory (Figure \ref{fig:Real_traj_pred}a).     Typically, the NN estimation  model fails to capture significantly drastic changes in the trajectory. Overall, the current end-to-end prediction model is robust to  out-of-distribution samples as well.

\section{Conclusion}

The current paper presents an end-to-end  estimator  that can take raw noisy sensor measurements and make robust future state predictions considering the upstream perceptual uncertainty. The  NN model uses deep ensembles and averages outputs over a batch of networks to provide the mean predicted path and associated uncertainty for each state. For perceptual uncertainty, the NN model approximates the characteristics of a  Bayes filter and estimates  the associated covariance. Further, the model also estimates the predictive uncertainty associated with future states to which the perceptual uncertainty is incorporated to obtain the total uncertainty.  Our results show that the incorporation of  sensing uncertainty into the prediction pipeline  enables the model to make robust downstream predictions. Overall, an ensemble model of 3 networks has been considered over a single network owing to  better  predictive uncertainty.  The performance metrics indicate that the mean predicted path for an ensemble model is closer to the ground truth compared to the MC dropout predictions.  Further, the coverage probability for an ensemble network is    higher  even with a smaller prediction interval width. Finally, the end-to-end prediction model showed robustness on out-of-distribution samples in quantifying both estimated future state and uncertainty. 

In the future, it will be interesting to consider non-parametric filters like particle filter (PF) that does not assume Gaussian distribution over the filter estimations and estimate whether the NN model performs better estimates by considering the PF.

\section{Biography Section}

	\begin{IEEEbiography}[{\includegraphics[width = 1 in, height = 1.25 in, clip, keepaspectratio]{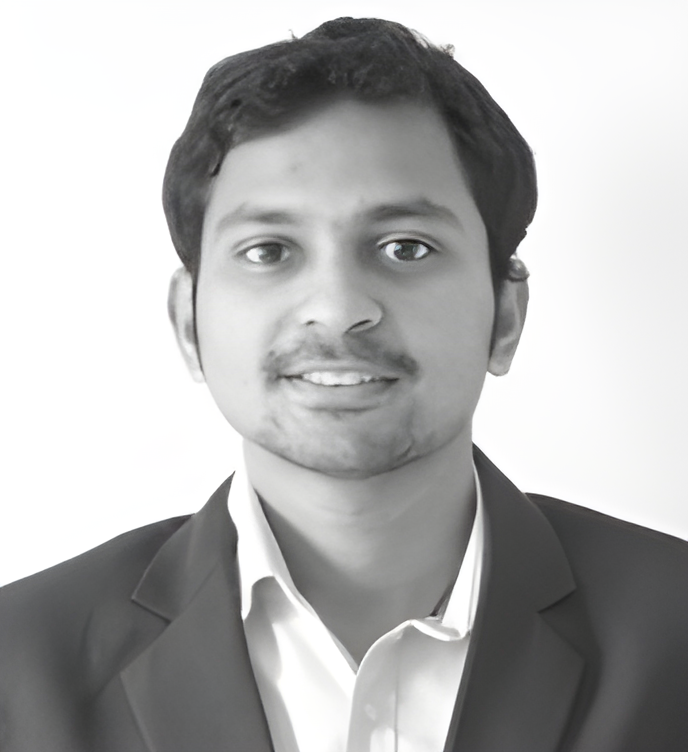}}]{Anshul Nayak}
	
		received his B.Tech  in mechanical engineering from NIT, Rourkela, India. He completed his Master's degree in Mechanical engineering at Virginia Tech and is currently pursuing his Ph.D at the  Autonomous Systems and Intelligent Machines (ASIM) lab at the same university. His research interests include cooperative planning and uncertainty estimation in prediction.
		
	\end{IEEEbiography}


\begin{IEEEbiography}[{\includegraphics[width = 1 in, height = 1.25 in, clip, keepaspectratio]{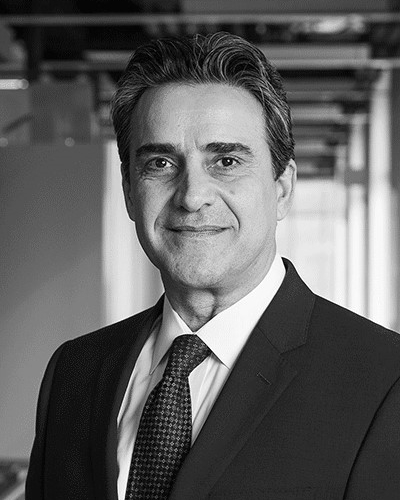}}]{Azim Eskandarian}
	
		has been a Professor and Head of the Mechanical Engineering Department at Virginia Tech since August 2015. He became the Nicholas and Rebecca Des Champs chaired Professor in April 2018. He also has a courtesy appointment as a Professor in the Electrical and Computer Engineering Department since 2021. He established the Autonomous Systems and Intelligent Machines laboratory at Virginia Tech and has conducted pioneering research in autonomous vehicles, human/driver cognition and vehicle interface, advanced driver assistance systems, and robotics. Before joining Virginia Tech, he was a Professor of Engineering and Applied Science at George Washington University (GWU) and the Founding Director of the Center for Intelligent Systems Research, from 1996 to 2015, the Director of the Transportation Safety and Security University Area of Excellence, from 2002 to 2015, and the Co-Founder of the National Crash Analysis Center in 1992 and its Director from 1998 to 2002 and 2013 to 2015. From 1989 to 1992, he was an Assistant Professor at Pennsylvania State University, York, PA, and an Engineer/Project Manager in the industry from 1983 to 1989. Dr. Eskandarian is a Fellow of ASME, a member of SAE, and a Senior Member of IEEE professional societies. He received SAE’s Vincent Bendix Automotive Electronics Engineering Award in 2021, IEEE ITS Society’s Outstanding Researcher Award in 2017, and GWU’s School of Engineering Outstanding Researcher Award in 2013.
			\end{IEEEbiography}

\begin{IEEEbiography}[{\includegraphics[width = 1 in, height = 1.25 in, clip, keepaspectratio]{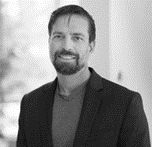}}]{Zachary Doerzaph}
	
	 is the Executive Director of the Virginia Tech Transportation Institute (VTTI), a global leader in transportation research.  Working alongside a talented team, Doerzaph focuses on creating a future of ubiquitous, safe, and effective mobility by conducting innovative and impactful research today.  Also, a faculty member within the Department of Biomedical Engineering and Mechanics at Virginia Tech, Doerzaph works with fellow faculty to provide experiential learning opportunities to prepare the next generation workforce. Doerzaph is known for innovative and extensive transportation research and leadership projects. His work focuses on maximizing performance at the interface of driver, vehicle, and infrastructure systems through the application of advanced technologies.
		
	\end{IEEEbiography}

\begin{IEEEbiography}[{\includegraphics[width = 1 in, height = 1.25 in, clip, keepaspectratio]{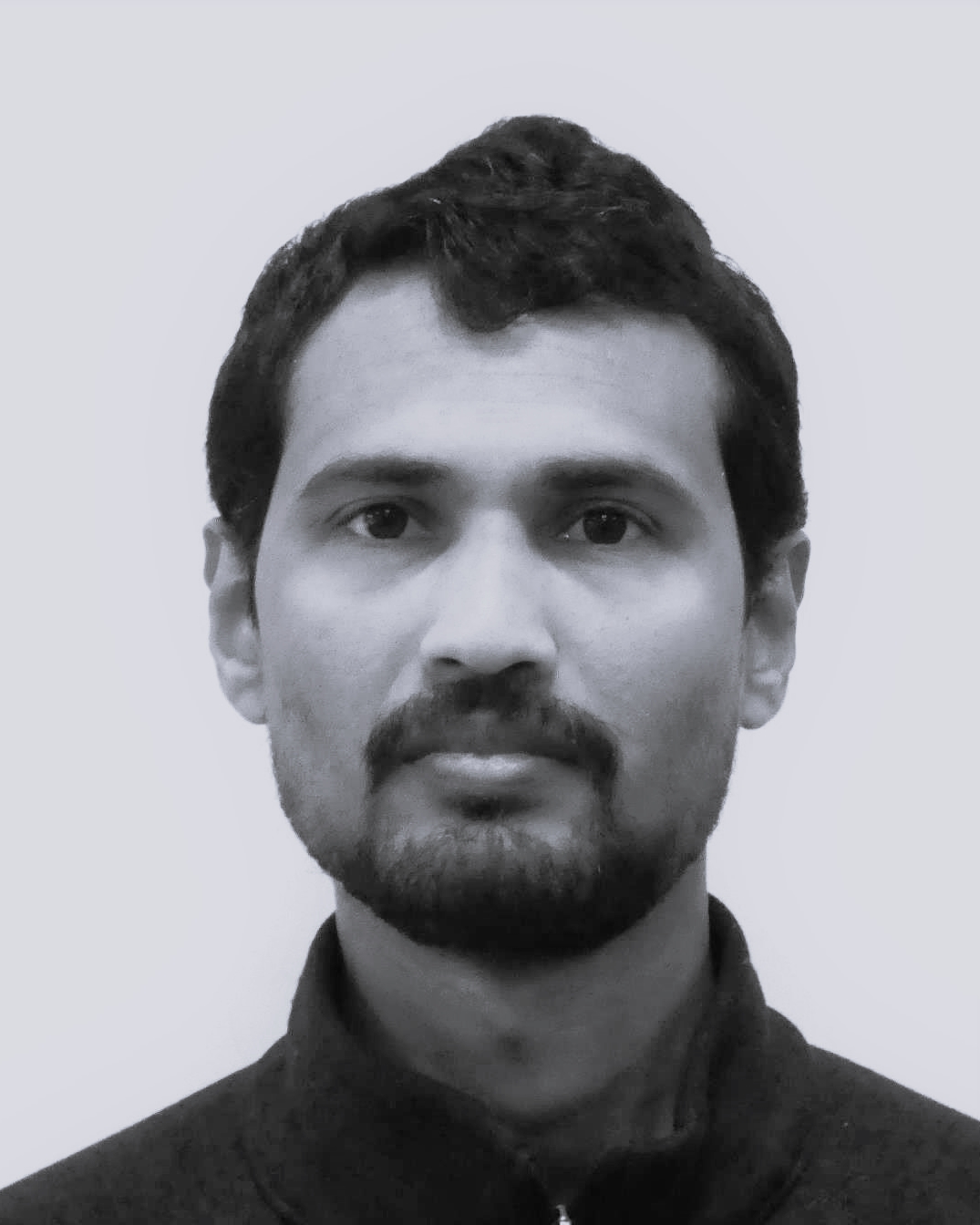}}]{Prasenjit Ghorai}
	
		received his B.Tech. degree from Maulana Abul Kalam Azad University of Technology (formerly West Bengal University of Technology, India) in electronics and instrumentation engineering, his M.Tech. degree in control \& instrumentation engineering from the University of Calcutta, and his Ph.D. degree in engineering from the National Institute of Technology (NIT) Agartala (in collaboration with Indian Institute of Technology, Guwahati, India). He was an Assistant Professor of Electronics and Instrumentation Engineering with NIT Agartala from 2011 to 2019. He is currently working as a senior research associate at the Autonomous Systems and Intelligent Machines Laboratory at Virginia Tech and conducts research on cooperative and connected autonomous vehicles.
		
	\end{IEEEbiography}

\vspace{11pt}

\vfill

\end{document}